\documentclass{article}

\usepackage[preprint]{neurips_data_2024}

\usepackage[utf8]{inputenc} 
\usepackage[T1]{fontenc}    
\usepackage{hyperref}       
\usepackage{url}            
\usepackage{booktabs}       
\usepackage{amsfonts}       
\usepackage{nicefrac}       
\usepackage{microtype}      
\usepackage{xcolor}         

\usepackage{graphicx}
\usepackage{enumitem}
\usepackage{bbding}
\usepackage{color}
\usepackage[noend]{algpseudocode}
\usepackage{algorithmicx,algorithm}
\usepackage{graphbox}
\usepackage{booktabs}
\usepackage{threeparttable}
\usepackage{multirow}
\usepackage{amsmath}
\usepackage{url}
\usepackage{tikz}
\usepackage{easyReview}
\usepackage{wrapfig,lipsum}
\usepackage{capt-of}
\def\halfcheckmark{\tikz\draw[scale=0.4,fill=black](0,.35) -- (.25,0) -- (1,.7) -- (.25,.15) -- cycle (0.75,0.2) -- (0.77,0.2)  -- (0.6,0.7) -- cycle;}

\usepackage[normalem]{ulem}

\title{\textsc{Concept} -- An Evaluation Protocol on Conversational Recommender Systems with System-centric\\ and User-centric Factors}

\author{
Chen Huang$^{\spadesuit}$, \quad
Peixin Qin$^{\spadesuit}$, \quad
Yang Deng$^{\heartsuit}$, \quad\\
\textbf{Wenqiang Lei}$^{\spadesuit}$\thanks{Corresponding author.}, \quad
\textbf{Jiancheng Lv}$^{\spadesuit}$, \quad
\textbf{Tat-Seng Chua}$^{\heartsuit}$
\\
${\spadesuit}$ Sichuan University \quad ${\heartsuit}$ National University of Singapore \\
\texttt{\{huangc.scu,qinpeixin.scu,dengyang17dydy\}@gmail.com}, \texttt{wenqianglei@scu.edu.cn}
}

\begin{document}

\maketitle

\begin{abstract}
The conversational recommendation system (CRS) has been criticized regarding its user experience in real-world scenarios, despite recent significant progress achieved in academia. Existing evaluation protocols for CRS may prioritize system-centric factors such as effectiveness and fluency in conversation while neglecting user-centric aspects. Thus, we propose a new and inclusive evaluation protocol, \textsc{Concept}, which integrates both system- and user-centric factors. We conceptualise three key characteristics in representing such factors and further divide them into six primary abilities. To implement \textsc{Concept}, we adopt a LLM-based user simulator and evaluator with scoring rubrics that are tailored for each primary ability. Our protocol, \textsc{Concept}, serves a dual purpose. First, it provides an overview of the {\it pros} and {\it cons} in current CRS models. Second, it pinpoints the problem of low usability in the "omnipotent" ChatGPT and offers a comprehensive reference guide for evaluating CRS, thereby setting the foundation for CRS improvement.
\end{abstract}

\section{Introduction}
\label{introthis}
The synergies between the conversation interface and recommendation system have given rise to a groundbreaking paradigm known as the Conversational Recommendation System (CRS)~\cite{sun2018conversational, gao2021advances}. It acts as a cooperative agent that engages in a natural language conversation with users and provides recommendations. Despite the great success achieved, CRS has been criticized regarding its user experience in real-world scenarios, lacking practical usability~\cite{jannach2020end, jannach2021survey}. This is partly due to the fact that current \textit{system-centric} evaluation protocols tend to prioritize assessing the characteristics of the CRS system {\it per se}, such as response diversity and fluency~\cite{ghazvininejad2018knowledge, wang2022barcor}, as well as the recommendation effectiveness and efficiency~\cite{wang-etal-2023-rethinking-evaluation, jin2019musicbot, warnestaal2005user}.
Such protocols often overlook \textit{user-centric} factors, which gauge how users engage with and perceive the social capabilities of the CRS. For instance, a CRS model may provide accurate recommendations and fluent conversations but at the same spread dishonest information. This can be misleading to users resulting in an unsatisfactory user experience.
Therefore, \textbf{it is imperative to consider incorporating both system- and user-centric factors in the evaluation protocol to develop a more user-friendly CRS system}.


\begin{figure*}
    \centering
    \setlength{\abovecaptionskip}{1pt}   
    \setlength{\belowcaptionskip}{1pt}
        \includegraphics[width=0.99\textwidth]{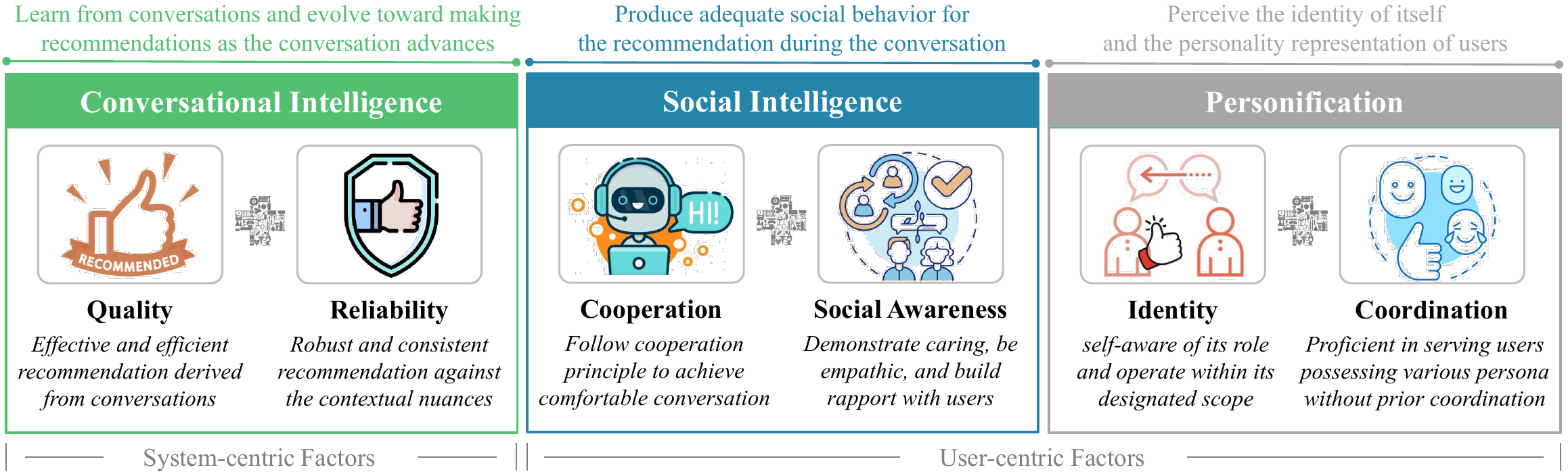}
    \caption{\textsc{Concept} integrates both system- and user-centric factors into three characteristics based on the previous taxonomy on human-AI interactions. Such characteristics are further divided into six primary abilities to enhance the inclusiveness in evaluations.}
    \label{fig:main}
    \vspace{-5mm}
\end{figure*}


To resolve this issue, we trace back the taxonomy that examines how the factors of conversational AI impact the user experience in human-AI interactions \cite{chaves2021should, 10.5555/236605}, and tailor a particular evaluation protocol for CRS.
We introduce \textbf{\textsc{Concept}}, an \underline{CO}mprehe\underline{N}sive \underline{C}RS \underline{E}valuation \underline{P}ro\underline{T}ocol. As framed in Figure \ref{fig:main}, \textsc{Concept} considers both system- and user-centric factors and conceptualizes them into three characteristics, which are further divided into six primary abilities. Such hierarchical factors are taken as inclusive and fine-grained evaluations. In addition, we present a practical implementation of \textsc{Concept} utilizing an LLM-based user simulator and evaluator, together with automated computational metrics. Specifically, the simulator mimics human social cognition and interacts with CRS to generate conversation data. Then, the evaluator assigns scores based on ability-specific scoring rubrics. This enables \textsc{Concept} to conduct labor-effective and inclusive evaluations.

By applying \textsc{Concept}, we can evaluate and analyze the strengths, weaknesses, and potential risks of off-the-shelf CRS models\footnote{We conduct evaluations using both humans and LLM, and found highly correlated results of the two.}.
A total of 6720 conversation data\footnote{\url{https://github.com/huangzichun/Concept4CRS}. To clarify, our contribution lies in the evaluation protocol, not the dataset. The dataset is generated dynamically alongside the execution of the protocol. } is recorded to collect the interactions between off-the-shelf CRS and simulated users who demonstrate different personas and preferences. Existing works agree on the insurmountable performance achieved by ChatGPT-based CRS. However, \textbf{current CRS models, even enhanced by "omnipotent" ChatGPT, still encounter numerous challenges, falling short of practical usability in particular}. 
1) They \textit{struggle to express genuine responses without hallucination or deceit}, often introducing non-existent items into conversations and present to users. 2) They \textit{lack of self-awareness of its identity, facing difficulties in producing both persuasive and honest explanations}, which is prominent in the ChatGPT-based CRS model, where explanations are highly convincing but frequently contain illusory details, misleading users in believing that these items align with their preferences. 3) They \textit{encounter issues in offering reliable recommendations} as they are sensitive to contextual nuances. Even slight alterations in user wording may result in entirely different recommendations. 4) They \textit{lack proficiency in catering to diverse users without prior coordination}, failing to dynamically adjust their behavior to align with each user's distinct personas. It is observed that the ChatGPT-based CRS model tends to employ deceptive tactics to persuade optimistic users to accept recommendations. This underscores the importance of aligning CRS with human values and advocating its ethical use. 
The main contributions as follows:
\begin{itemize}[leftmargin=*]
\setlength{\itemsep}{1pt}
\setlength{\parsep}{1pt}
\setlength{\parskip}{1pt}
    \item We pinpoint the fact that making a CRS admirable to users is primarily a social problem, instead of just a technical one. Social attributes are the key to the widespread acceptance of CRSs.
    \item We initiate the work on conceptualizing CRS’s characteristics in a comprehensive way, combining both system- and user-centric factors. 
    \item We propose a new evaluation protocol, called, \textsc{Concept}, which conceptualizes user expectations into six abilities, together with a scoring implementation.
    \item We evaluate and analyze the strengths, weaknesses, and potential risks of off-the-shelf CRS models in order to provide a fundamental perspective for people to make a reference for CRS evaluation.
\end{itemize}

\section{Related Work}
\textbf{CRS Evaluation}. We reckon that the success of CRS in practice may rely primarily on social characteristics, instead of just technical ones. However, existing evaluation protocols mainly focus on system-centric evaluation aspects, such as lexical diversity and perplexity of responses~\cite{ghazvininejad2018knowledge, chen2019towards}, or conversational fluency, relevance, and informativeness~\cite{wang2022towards, wang2022barcor}, human-CRS behavior alignment~\cite{yang2024behavior}, or recommendation effectiveness and efficiency~\cite{wang-etal-2023-rethinking-evaluation, jin2019musicbot, warnestaal2005user}. These aspects seem unable to reveal the inadequacy of CRS from user-centric perspectives. Hence, the fragmented evaluation protocols are underdeveloped to conduct an inclusive evaluation. Although some efforts have attempted to consider the user-centric characteristics of CRS, they rely on person-to-person conversation analysis and questionnaire interviews, lacking quantitative viewpoints and empirical evidence~\cite{10.1145/3631534, 10.1145/3472307.3484164, articledddd, 10.1145/3624989} in addition that they might have over-used system-centric characteristics\footnote{Refer to Table \ref{tab:rerer} in Appendix for better understanding our differences.}. In a nutshell, existing protocols may underestimate the problems of evaluating CRS~\cite{jannach2021survey, jannach2023evaluating}, and our \textsc{Concept} considers both system- and user-centric factors, aiming to provide an inclusive evaluation protocol for CRS.



\textbf{LLM as User Simulator}. Manual evaluation often requires the user to constantly interact with CRS, which are labor-expensive and generally only feasible in an industry lab~\cite{huang-etal-2023-reduce}. 
A recent study on CRS features interactive evaluation with LLM-based user simulation~\cite{wang-etal-2023-rethinking-evaluation}, experimentally demonstrating its effectiveness as a reliable alternative to humans. \textsc{Concept} also harnesses an LLM-based user simulator. Moreover, our simulator is equipped with the Theory of Mind~\cite{fischer2023reflective}, which enables the simulator to reflect on its predefined personas. 

\textbf{LLM as Evaluator}. A LLM-based evaluator could be seen as a combination of a LLM and its prompting strategy~\cite{zeng2023evaluating, cohen2023lm, chan2023chateval, wang2023chatgpt, liu2023gpteval}. 
Importantly, previous findings highlight that offering detailed scoring rubrics contributes to achieving consistent and aligned evaluations with human assessments~\cite{liu2023calibrating}. These findings are further supported by recent work on CRS~\cite{wang-etal-2023-rethinking-evaluation}, resulting in a dependable LLM-based evaluator as a viable alternative to human evaluators. Inspired by them, we involve ability-specific scoring rubrics to gain reliable and aligned evaluations. 

\section{\textsc{Concept}}
\label{skillset1}


Inspired by previous Interdisciplinary research on conversational AI \cite{chaves2021should, 10.5555/236605}, \textsc{Concept} consolidates both system- and user-centric factors into three characteristics and six specific abilities, as depicted in Figure \ref{fig:main}. 

\textbf{Factor 1: Recommendation Intelligence}. From a system-centric factors, this requires CRS to learn from conversations and evolve toward making recommendations as the conversation advances ~\cite{chen2019towards,ma2020cr,zhou2021crfr}. In this regard, our categorization encompasses two primary abilities:
\begin{itemize}[leftmargin=*]
    \item \textbf{Quality}. CRS should provide precise recommendations using minimal conversation turns, which are the crucial aspects that influence user satisfaction~\cite{10.1145/3624989, gao2021advances}. We emphasize the importance of the user acceptance rate, reflecting the practical effectiveness of the recommendations. 
    In our experiments, we utilize computational metrics for automatic evaluation, as seen in prior studies~\cite{wang-etal-2023-rethinking-evaluation, zhang2023variational, yu2023counterfactual}. These metrics encompass $Recall@k (k=1,10,25,50)$, recommendation success rate $(SR@k, k=3,5,10)$, user acceptance rate, and average turns ($AT$) needed to achieve successful recommendations.
    \item \textbf{Reliability}. CRS should deliver robust and consistent recommendations that account for contextual nuances. In practical situations, users with similar preferences may express themselves differently. It would diminish the user experience and pose a disruption for critical applications if inconsistent items were recommended for two similar user responses~\cite{tran2021recommender, oh2022rank}.
    In less critical cases, if two recommended items are inconsistent but align with user preferences, they are often viewed as diverse recommendations, creating opportunities for relevant but less popular items~\cite{10.1145/3459637.3481940}. 
    To evaluate reliability, we generate sets of user response pairs with similar meanings using ChatGPT paraphrasing. Given a user response pair [$u_1$, $u_2$], we define four computational metrics: the rate of \textit{Consistent Action}, indicating whether the CRS constantly provides recommendations based on $u_1$ and $u_2$; the rate of \textit{Consistent Recommendation}, which assesses if CRS recommends the same items given two user responses $u_1$ and $u_2$; and the rates of \textit{Diversity}, which evaluate whether the recommended items, even if inconsistent, align with user preferences. We refer to it as \textit{Sensitivity} when CRS provides inconsistent and inaccurate recommendations that do not align with user preferences.
\end{itemize}

\textbf{Factor 2: Social Intelligence}. It requires CRS to produce adequate social behavior for the recommendation during the conversations. 
As evidenced by the Media Equation Theory\footnote{Users tend to engage with the machine in a manner that mirrors person-to-person conversations.}~\cite{10.5555/236605, fogg2003computers}, users have high expectations for CRS to act cooperatively and be aware of user's social needs during the conversation, facilitating the design of CRS with perceived humanness~\cite{jacquet2018gricean, jacquet2019impact}. 
Our categorization encompasses two abilities.
\begin{itemize}[leftmargin=*]
    \item \textbf{Cooperation}. CRS should follow the cooperative principle to achieve comfortable conversations in common social situations. This is accomplished by adhering to the four "Maxims of Conversation"~\cite{grice1975logic, grice1989studies}, which form the basis for cooperative capability: 1) \underline{Manner}. CRS should respond in a manner that is easily understood and clearly expressed. 2) \underline{Sincerity}. CRS should communicate sincerely without deception, and ensure that its responses are backed by sufficient evidence. 3) \underline{Response Quality}. CRS should provide the necessary level of information for the conversation without overwhelming the user with unnecessary details. 4) \underline{Relevance}. CRS's responses should contribute to identifying user preferences and making recommendations. 
    \item \textbf{Social Awareness}. CRS must meet users' social expectations in practice, showing care, empathy, and establishing rapport with them~\cite{bjorkqvist2000social}. This facilitates the authenticity of CRS~\cite{neururer2018perceptions}. To achieve this, a recent study~\cite{hayati-etal-2020-inspired}  identified eight social strategies for building rapport with users, e.g., CRS could engage in self-disclosure and share its subjective opinion about a movie to establish a social connection with users.
\end{itemize}

We mainly resort to an LLM-based evaluator, except for sincerity evaluation (cf. Section \ref{evaluation}). For sincerity evaluation, we aim to utilize objective metrics: the ratio of non-existent items and deceptive tactics. For non-existent items, we tally how many CRS-recommended items do not exist in the dataset. As for the deceptive tactics, we focus on the items accepted by the user and track how many of these do not align with user preferences. If the user accepts these items, we consider the CRS to be employing deceptive tactics, leading the user to believe that the items meet his preferences. 

\begin{table*}[]
\centering
\setlength{\abovecaptionskip}{1pt}   
    \setlength{\belowcaptionskip}{1pt}
\resizebox{1.0\textwidth}{!}{%
\begin{tabular}{ll|l}
\toprule
\multicolumn{1}{l|}{\textbf{Abilities}} & \textbf{Descriptions} & \textbf{Evaluation metrics} \\ \midrule
\multicolumn{1}{l|}{\textbf{Quality}} & \begin{tabular}[c]{@{}l@{}}Provide precise recommendations \\ using minimal conversation turns\end{tabular} & \begin{tabular}[c]{@{}l@{}}Computational metrics \textit{using} the User Acceptance Rate\end{tabular} \\ \hline
\multicolumn{1}{l|}{\textbf{Reliability}} & \begin{tabular}[c]{@{}l@{}}Deliver robust and consistent recommendations \\ that account for contextual nuances\end{tabular} & \begin{tabular}[c]{@{}l@{}}Computational metrics \textit{using} the ratio of inconsistent recommendation \\and the ratio of recommendation sensitivity \end{tabular} \\ \midrule
\multicolumn{1}{l|}{\textbf{Cooperation}} & \begin{tabular}[c]{@{}l@{}}Follow cooperative principle \\to achieve comfortable conversations\end{tabular} & \begin{tabular}[c]{@{}l@{}}The average score of the Manner, Sincerity, \\Response Quality, and Relevance\end{tabular} \\ \hline
\multicolumn{1}{l|}{$\quad$1. Manner} & \begin{tabular}[c]{@{}l@{}}Response should be easily understood \\ and clearly expressed\end{tabular} & Ability-specific scoring \\ \hline
\multicolumn{1}{l|}{$\quad$2. Sincerity} & \begin{tabular}[c]{@{}l@{}}Communicate sincerely, \\without deception of pretense\end{tabular} & \begin{tabular}[c]{@{}l@{}}Computational metrics \textit{using} the ratio of deceptive tactics\\ and the ratio of non-existent items \\ 
\end{tabular} \\ \hline
\multicolumn{1}{l|}{$\quad$3. Response Quality} & \begin{tabular}[c]{@{}l@{}}Provide the necessary level of information \\ without unnecessary details\end{tabular}  & Ability-specific scoring\\ \hline
\multicolumn{1}{l|}{$\quad$4. Relevance} & \begin{tabular}[c]{@{}l@{}}Responses should contribute \\to making recommendations\end{tabular} & Ability-specific scoring \\ \hline
\multicolumn{1}{l|}{\textbf{Social Awareness}} & \begin{tabular}[c]{@{}l@{}}Meet user social expectations, \\establishing rapport with them\end{tabular} & Ability-specific scoring \\ \midrule
\multicolumn{1}{l|}{\textbf{Identity}} & \begin{tabular}[c]{@{}l@{}}Self-aware of its identity \\and operate within its designated scope\end{tabular} & \begin{tabular}[c]{@{}l@{}}Computational metrics \textit{using} Ratio of deceptive tactics\end{tabular} \\ \hline
\multicolumn{1}{l|}{\textbf{Coordination}} & \begin{tabular}[c]{@{}l@{}}Proficient in serving various and unknown users \\ without prior coordination\end{tabular} & \begin{tabular}[c]{@{}l@{}}Computational metrics
\textit{using} the range and mean of \\other ability-specific scores that are calculated among various users.
\end{tabular} \\ \bottomrule
\end{tabular}%
}
\caption{Summary of the evaluation taxonomy, descriptions of abilities, and evaluation metrics in \textsc{Concept}. LLM-based evaluator is used for ability-specific scoring, whereas computational metrics are used for the rest (cf. Table \ref{tab:summary_full} for details). We adjust the score to a scale of 1 to 5 when needed.}
\label{tab:summary}
\vspace{-4mm}
\end{table*}

\textbf{Factor 3: Personification}. Personification requires CRS to perceive the identity of itself and the personality representation of users. This involves CRS being self-aware of its role as a conversational recommendation system tailored for the general public.
\begin{itemize}[leftmargin=*]
    \item \textbf{Identity}. CRS should be self-aware of its identity and operate within its designated scope, differentiating itself from sales systems or other types. This ensures that CRS effectively carries out its identity by offering persuasive explanations to boost user acceptance~\cite{jannach2021survey, zhou-etal-2022-aligning}, avoiding resorting to misleading strategies (i.e., sales pitches with deceptive tactics). Otherwise, user trust and loyalty are hindered~\cite{gkika2014investigating}. Using misleading strategies also violates the maxim of sincerity (cf. Cooperation factor). We emphasize the need to integrate self-awareness into CRS to mitigate any deceptive behavior.
    We evaluate the identity by calculating the proportion of non-deceptive explanations. We also resort to the LLM-based evaluator to score the persuasiveness of recommendation explanations.
    \item \textbf{Coordination}. CRS should be proficient in serving users possessing various personas without prior coordination. Attributing personality to a conversational AI ensures that its behaviors stand in agreement with the users' expectations in a particular context \cite{chaves2021should}. This becomes particularly challenging for CRS, as it frequently serves users with diverse personas in real-world scenarios~\cite{thompson2004personalized}. As a result, CRS needs to exhibit various personalities and adapt its behavior to suit different users \cite{katayama2019situation, svikhnushina2021user, zhou2020design}. A crucial ability is to proficiently serve users without prior coordination. 
    We create simulations of users with different personas and evaluate the performance of the CRS across all the abilities mentioned. To assess coordination, we initially calculate the range\footnote{The range is more effective than the standard deviation in highlighting the variability of the CRS across different users.} and mean of the CRS's scores for each specific ability across different users. Then, we divide the range by the mean to determine the coordination score of the CRS for that specific ability. The overall coordination score is subsequently calculated as the average across all abilities.
\end{itemize}

\section{Experiment and Evaluation}
\subsection{Experimental Setup}
\label{setp}
\begin{wraptable}{r}{0.25\textwidth}
\centering
\setlength{\abovecaptionskip}{2pt}   
\setlength{\belowcaptionskip}{2pt}
\small
\resizebox{0.25\textwidth}{!}{%
\begin{tabular}{l|l}
\toprule
\textbf{Statistics}      & \textbf{Num} \\ \midrule
\# Conversations         & 6720         \\
Max Turns                & 10           \\
Avg. Turns               & 8.92         \\
Persona Types             & 12          \\ \bottomrule
\end{tabular}%
}
\caption{Data statistics.}
\label{tab:st}
\vspace{-3mm}
\end{wraptable}
\textbf{Dataset}. We utilize GPT-3.5-16K-turbo to create personas and play the roles of the LLM-based user simulator and evaluator. Following \citet{wang-etal-2023-rethinking-evaluation}, user preferences are defined using attributes from two benchmark datasets, i.e., Redial~\cite{li2018towards} and OpendialKG~\cite{moon-etal-2019-opendialkg}. We report the statistical summary of the generated conversation dataset in Table \ref{tab:st}.

\textbf{CRS Model}. Following~\citep{wang-etal-2023-rethinking-evaluation}, we present a comparative evaluation and analysis towards representative and SOTA CRS models, including \textit{KBRD}~\cite{chen2019towards}, \textit{BARCOR}~\cite{wang2022barcor}, \textit{UNICRS}~\cite{wang2022towards}, and \textit{CHATCRS}~\cite{wang-etal-2023-rethinking-evaluation}. 

\subsection{Evaluation Using \textsc{Concept}}
\label{evaluation}
Following the recent study on CRS~\cite{wang-etal-2023-rethinking-evaluation}, \textsc{Concept} resorts to an LLM-based user simulator and evaluator for cost-effective evaluation, together with computational metrics. We summary the taxonomy and evaluation metrics in Table \ref{tab:summary}. See Appendix \ref{sowef} for evaluation details.

\textbf{Evaluation Process}. The simulator creates simulations of users with diverse personas and preferences to interact with the CRS in producing conversation data. Afterwards, the evaluator assigns scores based on ability-specific scoring rubrics. Specifically, \textsc{Concept} considers the free-form chit-chat between the user simulator and CRS. To simulate real-world scenarios, the simulator has no access to its targeted items during the conversation. Any item that meets these preferences, such as having attributes completely consistent with or containing the simulator's preferences, is considered a successful recommendation. During the conversation, \textsc{Concept} allows the simulator to describe their preferences in their own words, as in real-world situations, users may not use the exact terms defined in the pre-defined preference values. Finally, the conversation will end if the simulator accepts recommendations or if the conversation reaches the maximum number of turns. Finally, \textsc{Concept} utilizes both the LLM-based evaluator and computational metrics to assess the abilities of CRS. Note that the evaluator is utilized when corresponding computational metrics are not available. 



\textbf{LLM-based User Simulator}. Our user simulator has unique personas and preferences. The personas are generated by prompting ChatGPT in a zero-shot manner, following~\citep{wang2023survey}, while the preferences are defined using attributes from two benchmark datasets (refer to Section \ref{setp}). In addition, \textsc{Concept} incorporates the Theory of Mind into our simulator to emulate human social cognition~\cite{fischer2023reflective}. This is achieved by prompting the simulator to first assess its current mental state before generating responses, eliciting reflection on its predefined personality traits and social interactions. See Appendix \ref{thised} for more implementation details on the simulator.

\textbf{LLM-based Evaluator}. Following previous studies~\cite{ye2023flask, wang2023large}, we employ the instance-wise evaluator. Building upon prior research~\cite{ wang-etal-2023-rethinking-evaluation, liu2023calibrating}, we prompt the evaluator with fine-grained scoring rubrics to eliminate the scoring bias. The evaluator assigns a score ranging from 1 to 5 to the conversation data using ability-specific rubrics, each accompanied by a corresponding description. For the generation of fine-grained rubrics, we follow the approach of previous works \cite{saha2023branch, li2023collaborative} and employ ChatGPT to produce a set of evaluation criterion, which are then refined by humans. See Appendix \ref{llmevaluat} for details.



\subsection{Evaluating off-the-shelf CRS models}
\label{evalua}

\begin{wrapfigure}{r}{0.36\textwidth}
    \centering
    \setlength{\abovecaptionskip}{1pt}   
    \setlength{\belowcaptionskip}{1pt}
    \includegraphics[width=0.35\textwidth]{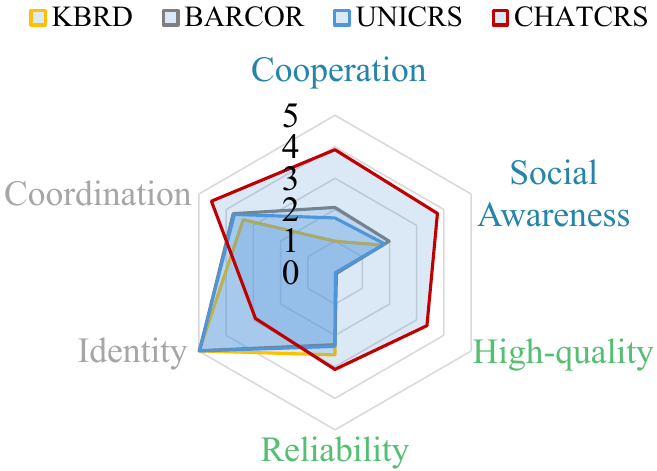}
    \caption{Evaluation overview.}
    \label{fig:radar_overall}
    \vspace{-7mm}
\end{wrapfigure}
Figure \ref{fig:radar_overall} presents an overview of the results across the six primary abilities, averaged across two benchmark datasets\footnote{For results on each benchmark, refer to Appendix \ref{moreandmore}.}. These results suggest that CHATCRS has made significant progress in cooperation, social awareness, and recommendation quality while losing advantages in identity. We will continue the current discussion in subsequent sections and provide further analysis.

\subsubsection{Recommendation-centric Evaluation}
\label{asdf3}

\begin{table*}[]
\centering
\setlength{\abovecaptionskip}{2pt}   
    \setlength{\belowcaptionskip}{2pt}
\resizebox{0.99\textwidth}{!}{%
\begin{tabular}{cl|cccc|cccc}
\toprule
\multicolumn{2}{c|}{\multirow{2}{*}{\textbf{Metrics}}} & \multicolumn{4}{c|}{\textbf{Redial}} & \multicolumn{4}{c}{\textbf{OpendialKG}} \\ \cline{3-10} 
\multicolumn{2}{c|}{} & \multicolumn{1}{l|}{\textbf{KBRD}} & \multicolumn{1}{l|}{\textbf{BARCOR}} & \multicolumn{1}{l|}{\textbf{UNICRS}} & \textbf{CHATCRS} & \multicolumn{1}{l|}{\textbf{KBRD}} & \multicolumn{1}{l|}{\textbf{BARCOR}} & \multicolumn{1}{l|}{\textbf{UNICRS}} & \textbf{CHATCRS} \\ \midrule
\multicolumn{1}{c|}{\multirow{8}{*}{\begin{tabular}[c]{@{}c@{}}Recommendation\\Module\\Perspective\end{tabular}}} & Recall@1 & 0.02&0.22&0.13&\textbf{0.41}&0.12&0.03&0.15&\textbf{0.37}\\ 
\multicolumn{1}{c|}{} & Recall@10 &0.23&1.37&1.09&\textbf{2.27}&0.98&0.94&1.28&\textbf{3.23} \\
\multicolumn{1}{c|}{} & Recall@25 & 0.57&3.23&2.44&\textbf{4.95}&1.94&2.07&3.06&\textbf{8.20}\\
\multicolumn{1}{c|}{} & Recall@50 & 1.13&5.69&4.58&\textbf{8.85}&3.53&3.43&5.81&\textbf{15.14} \\ \cline{2-10}
\multicolumn{1}{c|}{} & SR@3 & 3.95 & 31.36 & 14.04 & \textbf{37.72} & 4.69 & 1.82 & 9.90 & \textbf{31.12} \\ 
\multicolumn{1}{c|}{} & SR@5 & 4.39 & 35.53 & 15.68 & \textbf{40.90} & 14.19 & 3.52 & 17.45 & \textbf{37.24} \\ 
\multicolumn{1}{c|}{} & SR@10 & 4.50 & 39.47 & 18.20 & \textbf{46.60} & 16.02 & 7.29 & 29.30 & \textbf{46.48} \\ \cline{2-10}
\multicolumn{1}{c|}{} & AT ($\downarrow$) & 3.30 & 3.80 & 2.86 & \textbf{2.50} & 4.07 & 4.19 & 5.14 & \textbf{3.56} \\ \midrule
\multicolumn{1}{c|}{\multirow{4}{*}{\begin{tabular}[c]{@{}c@{}}Conversation\\ Module Perspective\end{tabular}}} & SR@3 & 20.18 & 27.52 & 35.20 & \textbf{52.63} & 6.51 & 17.71 & 14.58 & \textbf{26.30} \\ 
\multicolumn{1}{c|}{} & SR@5 & 24.34 & 39.47 & 38.27 & \textbf{58.55} & 10.68 & 24.22 & \textbf{26.69} & \textbf{36.33} \\ 
\multicolumn{1}{c|}{} & SR@10 & 29.39 & 50.66 & 43.42 & \textbf{62.39} & 12.37 & 35.16 & \textbf{45.31} & \textbf{44.40} \\ \cline{2-10}
\multicolumn{1}{c|}{} & AT ($\downarrow$) & \textbf{2.07} & 2.87 & 3.02 & 3.23 & 3.97 & 5.88 & 5.00 & \textbf{3.74} \\ \midrule
\multicolumn{1}{c|}{\multirow{2}{*}{User Perspective}} & Acceptance Rate & 0.33 & 1.43 & 0.33 & \textbf{70.83} & 0.39 & 0.65 & 0.26 & \textbf{64.32} \\ \cline{2-10}
\multicolumn{1}{c|}{} & AT ($\downarrow$) & 8.01 & 5.62 & 7.67 & \textbf{4.75} & 5.33 & 6.40 & 5.00 & \textbf{4.69} \\ \bottomrule
\end{tabular}%
}
\caption{Recommendation quality evaluation (\%) from three different perspectives. The average turn (AT) is calculated based on the corresponding conversation data with successful recommendations.}
\label{tab:main}
\vspace{-5mm}
\end{table*}

\textbf{CHATCRS stands out as the leading CRS model, delivering higher-quality recommendations}. As indicated in Table \ref{tab:main}, most CRS models have demonstrated improved success rates (SR) from the conversation module's perspective, indicating the effectiveness of optimizing conversation and recommendation modules simultaneously. 
However, we observed lower SR values for BARCOR on the Redial dataset compared to those from the recommendation module perspective, which is linked to the inclusion of non-existent items by its conversation module, such as "The Adventures of Milo and \underline{Ours}" (should be \textit{'Otis'}) and "The \underline{Prestigige}" (should be \textit{'Prestige'}).
Despite this, their recall\footnote{We only specifies user preferences, leading to lots of target items for users, resulting in a low recall value.} and SR values still fall short. In addition, the user acceptance rate of CRS models other than CHATCRS is notably low, possibly due to users' intolerance towards poor recommendations and the absence of persuasive recommendation explanations, which we will investigate on in further studies. Instead, CHATCRS stands out as the leading CRS model, with significant performance improvements, conforming to recent findings \cite{wang-etal-2023-rethinking-evaluation}. This may be attributed to two factors: 1) The text-embedding-ada-002 is much stronger in terms of translating conversation context and user preferences into embeddings, which helps enhance the recommendation module significantly. 2) CHATCRS are persuasive in convincing users to accept the recommendations. However, such a high user acceptance rate is largely based on deceptive tactics, a topic we will delve into later.

\begin{wrapfigure}{r}{0.53\textwidth}
    \centering
    \setlength{\abovecaptionskip}{5pt}   
    \setlength{\belowcaptionskip}{2pt}
    \includegraphics[width=0.5\textwidth]{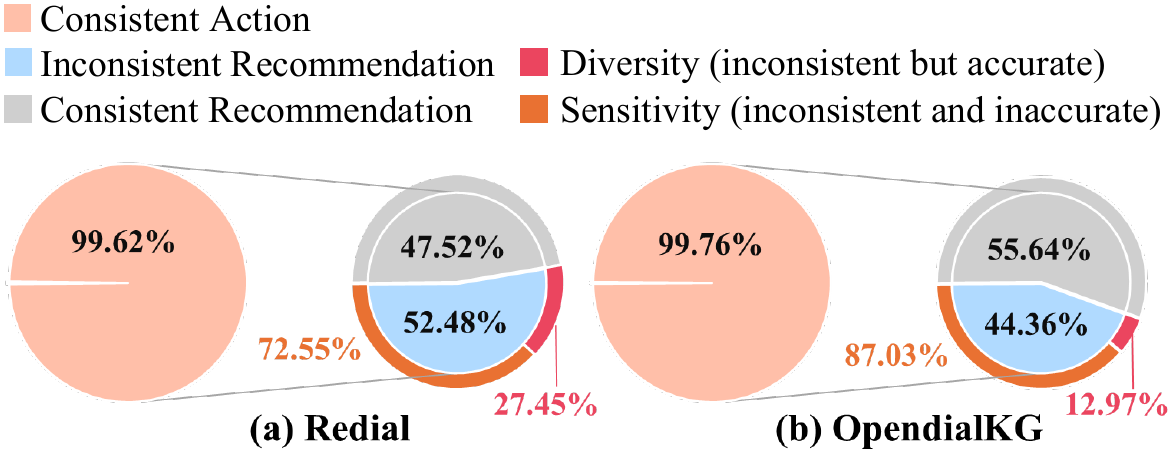}
    \caption{CHATCRS is sensitive to contextual nuance.}
    \label{fig:robust}
    \vspace{-4mm}
\end{wrapfigure}

\textbf{Current CRS strives to offer reliable recommendations due to its sensitivity to contextual nuance}.
We gauge the model reliability by evaluating how off-the-shelf CRS models handle user responses that are semantically similar but expressed differently. Results of CHATCRS are presented in Figure \ref{fig:robust}, which show promising action consistency of CHATCRS. It can make consistent recommendations when presented with two semantically similar user responses, achieving a consistency rate of over 99\%. However, the average recommendation consistency rate is much lower, at only 51.58\%, which means that 48.42\% of the time, if the user alters their wording slightly, CHATCRS will recommend entirely different items. To further explore its impact, we conducted a thorough analysis of the relationship between the attributes of the recommended items and user preferences. We discovered that only about 12\%-17\% of the recommended items align with user preferences, despite being inconsistent. This may increase chances for relevant but less popular items. As a consequence, the majority of recommendation results do not align with user preferences, demonstrating the sensitivity of current CRS to contextual nuances and the negative impact on user experience.

\subsubsection{Social-centric Evaluation}
\begin{wrapfigure}{r}{0.45\textwidth}
    \centering
    \setlength{\abovecaptionskip}{1pt}   
    \setlength{\belowcaptionskip}{1pt}
    \includegraphics[width=0.44\textwidth]{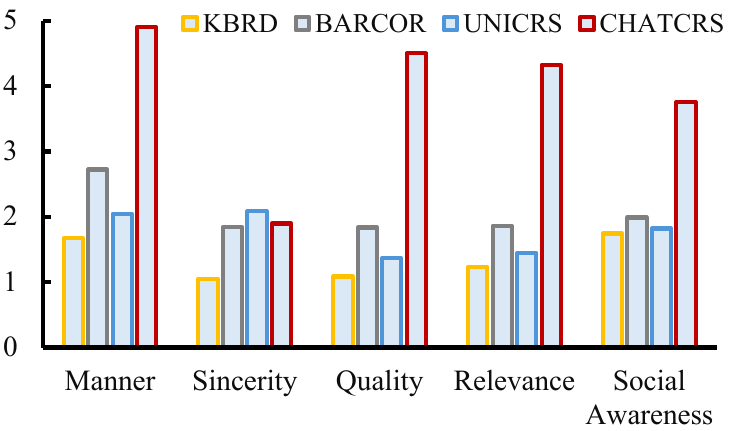}
    \caption{Evaluation on social-centric characteristics. CRS strives to self-express sincerely.}
    \label{fig:social}
\end{wrapfigure}

We break down cooperation ability into four maxims with an in-depth analysis, together with the social awareness evaluation. The results are depicted in Figure \ref{fig:social}. 

\textbf{CHATCRS manifests politeness and conversational traits in terms of manner, quality, relevance, and social awareness}. Different from other methods that lack a sense of topic or unable to maintain continuous conversations, CHATCRS has greatly improved its language abilities and social skills, which can be directly attributed to the impressive capabilities of ChatGPT in NLU, NLG, and empathy.
However, there is still room for improvement in terms of social awareness, even for CHATCRS. For example, there are instances where CHATCRS recommends items generated from a few rounds ago. As evidenced by previous studies~\cite{portela2017new}, keeping track of the conversational history is reported as an empathic behavior, leading to the rise of affection. This calls for the investigation on a more social-aware CRS.

\begin{figure}
\raisebox{1\height}{%
  \begin{minipage}[t]{0.45\linewidth}
    \centering
    \resizebox{0.8\textwidth}{!}{%
	\begin{tabular}{l|ccc}
	\toprule
	\textbf{CRS} & \textbf{Redial} & \textbf{OpendialKG} & \textbf{Avg.} \\ \midrule
	KBRD & 1.02 & 1.00 & 1.01\\
	BARCOR & 1.55 & 1.25 & 1.40\\
	UNICRS & 1.08 & 1.06 & 1.07\\
	CHATCRS & \textbf{4.66} & \textbf{4.48} & \textbf{4.57}\\ \bottomrule 
\end{tabular}%
 }
	\captionof{table}{Results of persuasiveness scores. CHATCRS are highly persuasive.}
	\label{tab:my-expl}
  \end{minipage}%
  }
  \begin{minipage}[t]{0.55\linewidth}
    \centering
    \setlength{\abovecaptionskip}{1pt}   
\setlength{\belowcaptionskip}{1pt}
    \includegraphics[width=0.9\textwidth]{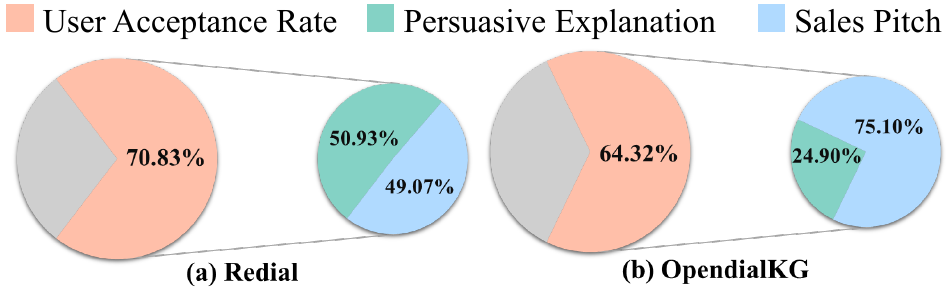}
    \caption{CHATCRS offers persuasive yet dishonest explanations.}
    \label{fig:sales}
  \end{minipage}
  \vspace{-3mm}
\end{figure}

\textbf{Current CRS struggle to express genuine responses without hallucination or deceit}. According to Fig. \ref{fig:social}, sincerity scores of CRS models are far from satisfactory, which can be attributed to two factors: the recommendation hallucination and dishonest explanations. Regarding the hallucination, CRS tends to introduce non-existent items into conversations and promote them to users. Even in the case of CHATCRS, its responses still contain 5.18\% of non-existent items on Redial and 7.42\% on OpendialKG. As for dishonest explanations, CRS tends to utilize persuasive language to mislead users into accepting recommendations by providing false explanations in movie plots and attributes. This problem is even more severe in CHATCRS, where, on average, approximately 62.09\% of the explanations do not meet the requirements of sincerity (cf. Section \ref{asdf55}, Fig. \ref{fig:sales}). 



\subsubsection{Personification-centric Evaluation}
\label{asdf55}

\textbf{Lacking self-awareness, CRS models often offers persuasive yet dishonest explanations}. As in Table \ref{tab:my-expl}, most CRS models aim for being persuasive in their recommendations. Explanations may sound lackluster, insipid, or feeble, simply stating that "\textit{this is a good movie}," or making absurd claims such as "\textit{Jumanji (2017) is about a man who is a human!}". These recommendations become less acceptable for users. In contrast, CHATCRS consistently furnishes explanations for its recommendations, employing text-based logical reasoning to enhance comprehensibility. This accounts for its high user acceptance rate. However, as in Figure \ref{fig:sales}, although highly convincing, the explanations from CHATCRS often include illusory details, leading users to mistakenly believe that these items align with their preferences. Taking OpendialKG for example, 75.10\% of accepted items do not align with user preferences. Given that the majority of CRS models employ reinforcement learning, or RLHF, this could cause problems such as reward hacking and misspecification~\cite{pan2022effects}. Such problems can drive CRS to acquire deceitful behavior. Considering the above disadvantages, the urgency is to create an identity-aware CRS that can deliver both persuasive and honest explanations.

\begin{figure*}[t]
    \centering
    \setlength{\abovecaptionskip}{1pt}   
    \setlength{\belowcaptionskip}{1pt}
    \includegraphics[width=0.98\textwidth]{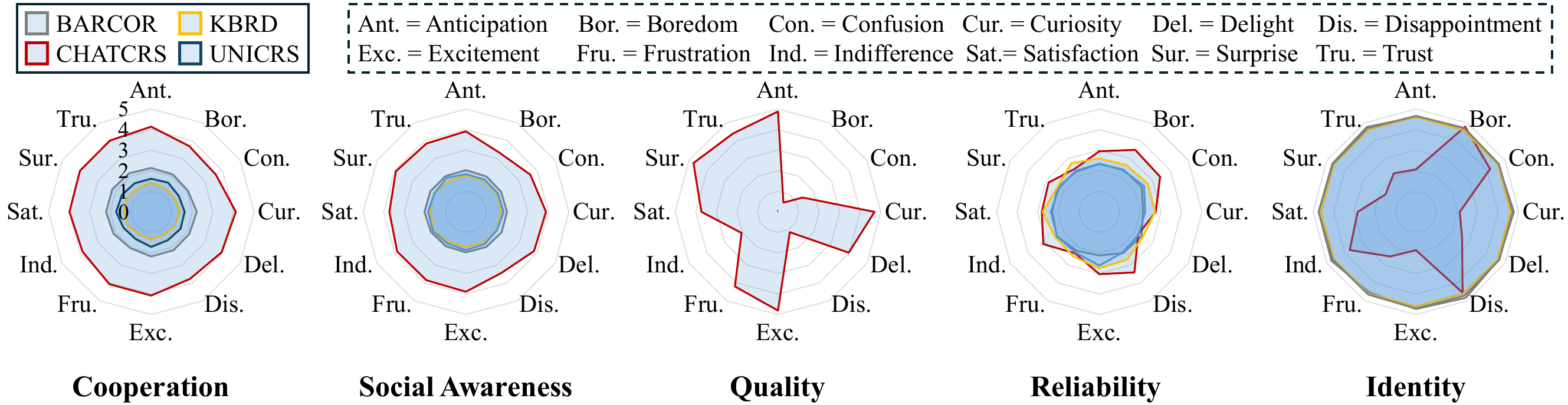}
    \caption{Coordination evaluation on users with various personas (marked with dashed boxes). CHATCRS is hard to cater to diverse users, while others consistently show low performance levels. 
    Quality score of models besides CHATCRS are extremely low (cf. \textit{Acceptance Rate} row in Table \ref{tab:main}).
    }
    \label{fig:personacoor}
    \vspace{-5mm}
\end{figure*}

\textbf{Lacking proficiency in prior coordination, CRS models often fails to cater different user needs}. 
As in Figure \ref{fig:personacoor}, most CRS models, except for CHATCRS, show poor performance in sensing the variation of users. In contrast, CHATCRS outperforms others in general and is more sensitive to different users, providing high-quality recommendations. In particular, CHATCRS can properly deal with users' negative emotions, such as bored, confused, or disappointed. Interestingly, according to Quality score, these people tend to show low acceptance of recommendations, i.e. they possess a higher acceptance threshold for recommendation quality. Such observations drives us in developing a CRS that can dynamically adjust its recommendation strategy to align with the unique personas of each user. Apart from that, we also examine the interaction behavior of CHATCRS with different users. For instance, 
according to the Identity score,
CHATCRS adopts sales pitches with deceptive tactics to persuade optimistic users to accept recommendations. However, for pessimistic users, CHATCRS tends to provide persuasive and honest recommendation explanations. This somewhat reveals a bias in CHATCRS's recommendation strategy towards different user groups, which is a flaw that needs to be rectified in future work.

\subsection{Reliability of \textsc{Concept}} 
\label{rela}
\begin{wrapfigure}{r}{0.6\textwidth}
    \centering
    \setlength{\abovecaptionskip}{1pt}   
    \setlength{\belowcaptionskip}{1pt}
    \includegraphics[width=0.59\textwidth]{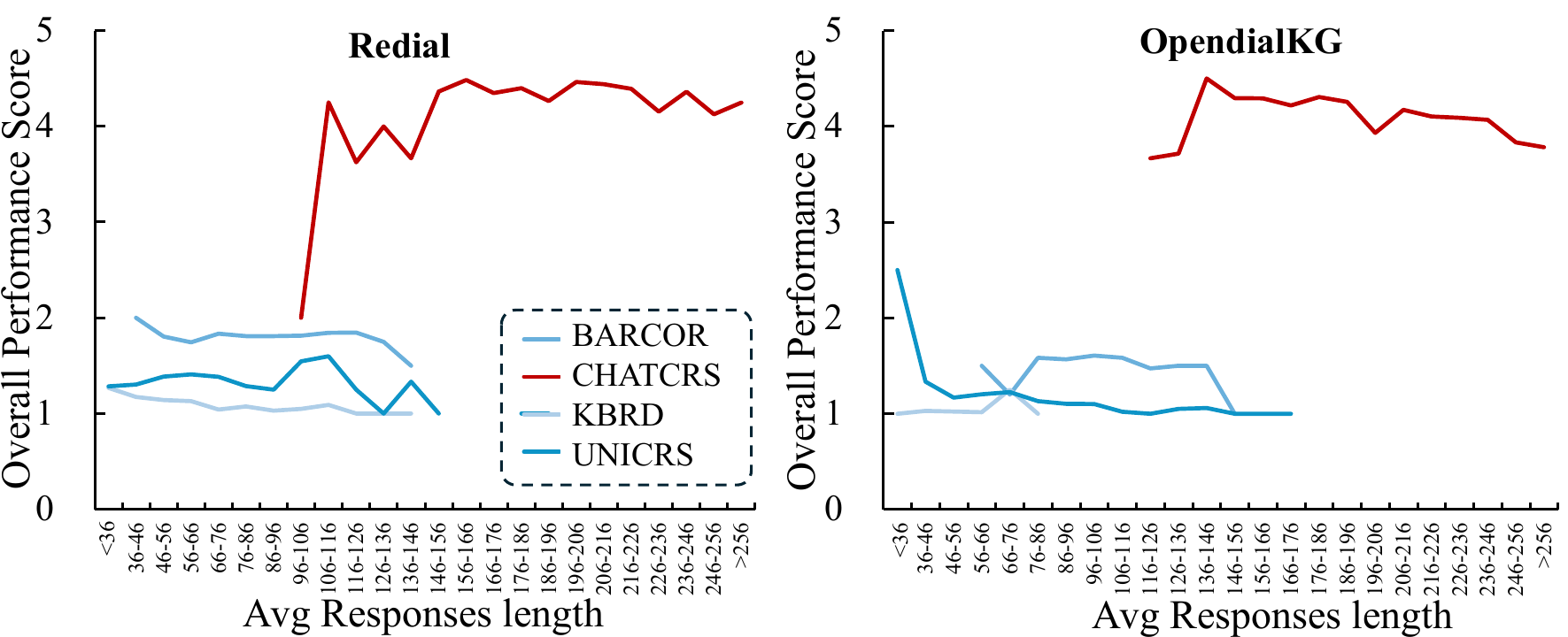}
    \caption{Length bias evaluation.}
    \label{fig:len}
    \vspace{-5mm}
\end{wrapfigure}
\textbf{Replicability Analysis}. We have established fixed values for the temperature and seed parameters to guarantee replicability of our LLM-based simulator and evaluator.

\textbf{Bias Analysis}. 
We build upon previous research~\cite{ye2023flask, wang-etal-2023-rethinking-evaluation, liu2023calibrating} and minimize potential biases by employing detailed scoring rubrics. In this study, we examine the evaluation biases\footnote{\textsc{Concept} does not entail making decisions based on a specific group of candidates, thus no position bias.}, and demonstrate that \textsc{Concept} scoring is unbiased and aligns with the results of human evaluations, consistent with earlier findings. 
\begin{itemize}[leftmargin=*]
    \item \textit{Length bias}\footnote{LLMs have a preference for longer responses~\cite{wu2023style}.}. As shown in Figure \ref{fig:len}, we require CHATGPT to provide an overall performance score (\textit{Y-axis}, Overall Performance Score) based on all the ability-specific scores and then plot the relationship between various CRS reply lengths (\textit{X-axis}, Avg. Response Length) and the scores. Our findings show that \textsc{Concept} remains unaffected by the length bias. CHATCRS tends to produce lengthier responses, but this does not imply that longer responses will yield higher scores. 
    \item \textit{Self-enhancement bias}\footnote{Bias of CHATGPT to favor high scores for its generated content \cite{wang2023large}.}. Our human evaluation demonstrates that the \textsc{Concept} evaluation results are consistent and align with human assessments  (cf. Appendix \ref{human} for more details). Specifically, the human evaluation results and the LLM-based evaluation results are closely related, with a correlation coefficient of 61.24\% and Krippendorff’s alpha of 53.10\%. This indicates the reliability of our LLM-based evaluation.
\end{itemize}

\section{Conclusion}
We regard CRS as a social issue, instead of just a technical problem, considering its major goal of persuading users to accept the recommendations. Hence, we revisit current evaluation protocols and pioneer works to provide a comprehensive survey on understanding the impact of CRS’s characteristics in terms of both pros and cons. We then come up with a new evaluation protocol-- \textsc{Concept}, which considers both system- and user-centric factors for addressing personalized user needs in particular, with the hope of enhancing user experience in CRS. Such factors include three key characteristics, which are further divided into six primary abilities. In addition, we suggest that even enhanced by "omnipotent" LLMs, CRS lacks practical usability. We pinpoint several significant limitations of current CRS models, including the incapacity to express itself sincerely, a lack of identity-awareness, low reliability towards contextual nuance, and poor coordination for diverse users. 
With \textsc{Concept}, we provide an overview for researchers as a reference guidance for evaluating CRS and laying the foundation for CRS enhancement.

\clearpage

\bibliographystyle{plainnat}
\bibliography{custom}
\clearpage
\appendix
\section{Limitation Analysis of Our Work}
\label{limitation}
The use of LLM-based user simulators and evaluators is a double-edged sword. It is a labor-intensive and effective approach, but may suffer from weak robustness by its nature. Although we have adopted some strategies to improve the robustness as much as possible following previous works, the robustness may still be hampered due to the potential uncertainty of prompt engineering. Additionally, another aspect that could improve the \textsc{Concept} is to generate more conversation data, run the LLM-based evaluation many times with different seeds, and report more statistically significant results. However, budgeting is always a factor to consider. In this case, it is important to propose a user simulator and evaluator based on an open-source small model that has similar capabilities to ChatGPT. Finally, our current work does not evaluate the attribute-based CRS \cite{lei2020conversational}, as this type of research often ignores the ability to engage in smooth conversations and instead focuses solely on accurately providing recommendations to users within minimal conversation turns. In this case, evaluating attribute-based CRS models seems unfair. We highlight the importance of combining attribute-based and dialog-based CRS studies to create a more holistic CRS, taking into account its practical usability.

\begin{wraptable}{r}{0.4\textwidth}
\centering
\resizebox{0.4\textwidth}{!}{%
\begin{tabular}{l|llll|llll}
\toprule
 & \multicolumn{4}{c|}{\textbf{Redial}} & \multicolumn{4}{c}{\textbf{OpendialKG}} \\
\multirow{-2}{*}{\textbf{User Types}} & \multicolumn{1}{c}{{\textbf{B}}} & \multicolumn{1}{c}{{\textbf{C}}} & \multicolumn{1}{c}{{\textbf{K}}} & \multicolumn{1}{c|}{{\textbf{U}}} & \multicolumn{1}{c}{{\textbf{B}}} & \multicolumn{1}{c}{{\textbf{C}}} & \multicolumn{1}{c}{{\textbf{K}}} & \multicolumn{1}{c}{{\textbf{U}}} \\ \midrule
\textbf{Anticipation } & \textbf{76} & \textbf{76} & \textbf{76} & \textbf{76} & \textbf{64} & \textbf{64} & \textbf{64} & \textbf{64} \\
Age group=Adults & 19 & 19 & 19 & 19 & 16 & 16 & 16 & 16 \\
Age group=Children & 19 & 19 & 19 & 19 & 16 & 16 & 16 & 16 \\
Age group=Seniors & 19 & 19 & 19 & 19 & 16 & 16 & 16 & 16 \\
Age group=Teens & 19 & 19 & 19 & 19 & 16 & 16 & 16 & 16 \\ \hline
\textbf{Boredom } & \textbf{76} & \textbf{76} & \textbf{76} & \textbf{76} & \textbf{64} & \textbf{64} & \textbf{64} & \textbf{64} \\
Age group=Adults & 19 & 19 & 19 & 19 & 16 & 16 & 16 & 16 \\
Age group=Children & 19 & 19 & 19 & 19 & 16 & 16 & 16 & 16 \\
Age group=Seniors & 19 & 19 & 19 & 19 & 16 & 16 & 16 & 16 \\
Age group=Teens & 19 & 19 & 19 & 19 & 16 & 16 & 16 & 16 \\ \hline
\textbf{Confusion } & \textbf{76} & \textbf{76} & \textbf{76} & \textbf{76} & \textbf{64} & \textbf{64} & \textbf{64} & \textbf{64} \\
Age group=Adults & 19 & 19 & 19 & 19 & 16 & 16 & 16 & 16 \\
Age group=Children & 19 & 19 & 19 & 19 & 16 & 16 & 16 & 16 \\
Age group=Seniors & 19 & 19 & 19 & 19 & 16 & 16 & 16 & 16 \\
Age group=Teens & 19 & 19 & 19 & 19 & 16 & 16 & 16 & 16 \\ \hline
\textbf{Curiosity } & \textbf{76} & \textbf{76} & \textbf{76} & \textbf{76} & \textbf{64} & \textbf{64} & \textbf{64} & \textbf{64} \\
Age group=Adults & 19 & 19 & 19 & 19 & 16 & 16 & 16 & 16 \\
Age group=Children & 19 & 19 & 19 & 19 & 16 & 16 & 16 & 16 \\
Age group=Seniors & 19 & 19 & 19 & 19 & 16 & 16 & 16 & 16 \\
Age group=Teens & 19 & 19 & 19 & 19 & 16 & 16 & 16 & 16 \\ \hline
\textbf{Delight } & \textbf{76} & \textbf{76} & \textbf{76} & \textbf{76} & \textbf{64} & \textbf{64} & \textbf{64} & \textbf{64} \\
Age group=Adults & 19 & 19 & 19 & 19 & 16 & 16 & 16 & 16 \\
Age group=Children & 19 & 19 & 19 & 19 & 16 & 16 & 16 & 16 \\
Age group=Seniors & 19 & 19 & 19 & 19 & 16 & 16 & 16 & 16 \\
Age group=Teens & 19 & 19 & 19 & 19 & 16 & 16 & 16 & 16 \\ \hline
\textbf{Disappointment } & \textbf{76} & \textbf{76} & \textbf{76} & \textbf{76} & \textbf{64} & \textbf{64} & \textbf{64} & \textbf{64} \\
Age group=Adults & 19 & 19 & 19 & 19 & 16 & 16 & 16 & 16 \\
Age group=Children & 19 & 19 & 19 & 19 & 16 & 16 & 16 & 16 \\
Age group=Seniors & 19 & 19 & 19 & 19 & 16 & 16 & 16 & 16 \\
Age group=Teens & 19 & 19 & 19 & 19 & 16 & 16 & 16 & 16 \\ \hline
\textbf{Excitement } & \textbf{76} & \textbf{76} & \textbf{76} & \textbf{76} & \textbf{64} & \textbf{64} & \textbf{64} & \textbf{64} \\
Age group=Adults & 19 & 19 & 19 & 19 & 16 & 16 & 16 & 16 \\
Age group=Children & 19 & 19 & 19 & 19 & 16 & 16 & 16 & 16 \\
Age group=Seniors & 19 & 19 & 19 & 19 & 16 & 16 & 16 & 16 \\
Age group=Teens & 19 & 19 & 19 & 19 & 16 & 16 & 16 & 16 \\ \hline
\textbf{Frustration } & \textbf{76} & \textbf{76} & \textbf{76} & \textbf{76} & \textbf{64} & \textbf{64} & \textbf{64} & \textbf{64} \\
Age group=Adults & 19 & 19 & 19 & 19 & 16 & 16 & 16 & 16 \\
Age group=Children & 19 & 19 & 19 & 19 & 16 & 16 & 16 & 16 \\
Age group=Seniors & 19 & 19 & 19 & 19 & 16 & 16 & 16 & 16 \\
Age group=Teens & 19 & 19 & 19 & 19 & 16 & 16 & 16 & 16 \\ \hline
\textbf{Indifference } & \textbf{76} & \textbf{76} & \textbf{76} & \textbf{76} & \textbf{64} & \textbf{64} & \textbf{64} & \textbf{64} \\
Age group=Adults & 19 & 19 & 19 & 19 & 16 & 16 & 16 & 16 \\
Age group=Children & 19 & 19 & 19 & 19 & 16 & 16 & 16 & 16 \\
Age group=Seniors & 19 & 19 & 19 & 19 & 16 & 16 & 16 & 16 \\
Age group=Teens & 19 & 19 & 19 & 19 & 16 & 16 & 16 & 16 \\ \hline
\textbf{Surprise } & \textbf{76} & \textbf{76} & \textbf{76} & \textbf{76} & \textbf{64} & \textbf{64} & \textbf{64} & \textbf{64} \\
Age group=Adults & 19 & 19 & 19 & 19 & 16 & 16 & 16 & 16 \\
Age group=Children & 19 & 19 & 19 & 19 & 16 & 16 & 16 & 16 \\
Age group=Seniors & 19 & 19 & 19 & 19 & 16 & 16 & 16 & 16 \\
Age group=Teens & 19 & 19 & 19 & 19 & 16 & 16 & 16 & 16 \\ \hline
\textbf{Trust } & \textbf{76} & \textbf{76} & \textbf{76} & \textbf{76} & \textbf{64} & \textbf{64} & \textbf{64} & \textbf{64} \\
Age group=Adults & 19 & 19 & 19 & 19 & 16 & 16 & 16 & 16 \\
Age group=Children & 19 & 19 & 19 & 19 & 16 & 16 & 16 & 16 \\
Age group=Seniors & 19 & 19 & 19 & 19 & 16 & 16 & 16 & 16 \\
Age group=Teens & 19 & 19 & 19 & 19 & 16 & 16 & 16 & 16 \\ \hline
\textbf{Satisfaction } & \textbf{76} & \textbf{76} & \textbf{76} & \textbf{76} & \textbf{64} & \textbf{64} & \textbf{64} & \textbf{64} \\
Age group=Adults & 19 & 19 & 19 & 19 & 16 & 16 & 16 & 16 \\
Age group=Children & 19 & 19 & 19 & 19 & 16 & 16 & 16 & 16 \\
Age group=Seniors & 19 & 19 & 19 & 19 & 16 & 16 & 16 & 16 \\
Age group=Teens & 19 & 19 & 19 & 19 & 16 & 16 & 16 & 16 \\
\midrule
\textbf{In Total} & \textbf{912} & \textbf{912} & \textbf{912} & \textbf{912} & \textbf{768} & \textbf{768} & \textbf{768} & \textbf{768} \\ \bottomrule
\end{tabular}%
}
\caption{{\small Number of conversational attributes. 'B' stands for BARCOR, 'C' for CHATCRS, 'K' for KBRD, 'U' for UNICRS.}}
\label{tab:data}
\end{wraptable}
\section{Implementation Details}
\label{detailsin}
We conduct all our experiments using a single Nvidia RTX A6000, and we implement our codes in PyTorch. The core of our code framework is built upon open-source code from the latest research on CRS\footnote{\url{https://github.com/RUCAIBox/iEvaLM-CRS}}~\cite{wang-etal-2023-rethinking-evaluation}. In order to guarantee replicability, we have established fixed values for the Temperature and Seed parameters of ChatGPT (i.e., GPT-3.5-16K-turbo), setting the Temperature to 0 and the Seed to 42.

\subsection{Implementation of CRS Models}
We evaluated all open-source CRS models in its code using their checkpoints, including the KBRD~\cite{chen2019towards}, BARCOR~\cite{wang2022barcor}, UNICRS~\cite{wang2022towards}, and CHATGPT-based model~\cite{wang-etal-2023-rethinking-evaluation} which represents the current SOTA, incorporating \textit{text-embedding-ada-002}~\cite{neelakantan2022text} for modeling the semantic embeddings. 
\begin{itemize}[leftmargin=*]
    \item \textbf{KBRD}~\cite{chen2019towards} bridges the recommendation module and the Transformer-based conversation module via knowledge propagation.
    \item \textbf{BARCOR}~\cite{wang2022barcor} is a unified framework based on BART~\cite{lewis2020bart}, which implements the recommendation and response generation tasks in a single model.
    \item \textbf{UNICRS}~\cite{wang2022towards} is a unified framework based on DialoGPT~\cite{zhang2020dialogpt}, with a semantic fusion module to enhance the semantic association between conversation history and knowledge graphs.
    \item \textbf{CHATCRS}~\cite{wang-etal-2023-rethinking-evaluation} is the SOTA CRS model, which incorporates ChatGPT for the conversation module and the text-embedding-ada-002~\cite{neelakantan2022text} to enhance the recommendation module. 
\end{itemize}
See the prompts in the original paper for the implementation of CHATGPT-based model. Note that we feed the top 5 items from the recommendation module into the prompts of ChatGPT for re-ranking and generating responses. Also, the UNICRS utilizes DialoGPT-small as the backbone, while the BARCOR utilizes BART-base with a 2-layer encoder and decoder, following \citet{wang-etal-2023-rethinking-evaluation}. 


\subsection{Implementation of \textsc{Concept}}
\label{sowef}
\textsc{Concept} resorts to an LLM-based user simulator and evaluator for cost-effective evaluation, together with fine-grained ability-specific scoring rubrics. Table \ref{tab:data} reveals the statistics of our generated conversation dataset.

\subsubsection{Evaluation Process}
We used user simulators to interact and converse with different CRSs to produce a conversation dataset. In the experiments, the simulator simulated 12 different personas and 4 different age groups using ChatGPT. Specifically, \textsc{Concept} considers the free-form chit-chat between the user simulator and CRS. To simulate real-world scenarios, the simulator has no access to its targeted items during the conversation. Any item that meets these preferences, such as having attributes completely consistent with or containing the simulator's preferences, is considered a successful recommendation. During the conversation, \textsc{Concept} allows the simulator to describe their preferences in their own words, as in real-world situations, users may not use the exact terms defined in the pre-defined preference values. During the conversation between CRS and the user, we recorded the recommendation results of each round of the recommendation system and the results recommended by the conversational agent, in order to evaluate the success of the recommendations from different perspectives. 
Finally, the conversation will end if the simulator accepts recommendations or if the conversation reaches the maximum number of turns.
If the user chooses to accept the recommendation from the conversational agent, the user is required to add '[END]' at the end of his response to indicate the end of the conversation. These diverse users generated a total of 6720 conversations, which were used to evaluate the performance of different CRS in more realistic scenarios.

Afterward, \textsc{Concept} utilizes both the LLM-based evaluator and computational metrics to assess the abilities of CRS, as outlined in Table \ref{tab:summary}. Note that the evaluator is utilized when corresponding computational metrics are not available. Specifically, we employ the evaluator to assess the cooperation ability, social awareness, and persuasiveness of recommendation explanations of CRS. This involves prompting the evaluator with fine-grained, ability-specific scoring rubrics \cite{saha2023branchsolvemerge}, which are generated by the LLM and then refined by humans. For more details, refer to Appendix \ref{llmevaluat} and Appendix \ref{automa}.

\subsubsection{LLM-based User Simulator}
\label{thised}
Our user simulator has unique personas and preferences. Specifically, the personas are generated by prompting ChatGPT in a zero-shot manner, following~\citep{wang2023survey}, while the preferences are defined using attributes from two benchmark datasets, Redial and OpendialKG. 

\textbf{Personas}. We generate a persona list as the starting point by prompting ChatGPT in a zero-shot manner. Inspired by~\cite{wang2023survey}, we first instruct the ChatGPT to return 20 distinct personas and their corresponding descriptions, such as curiosity, and then filter out duplicates. We utilize the generated persona list, so it is highly likely that ChatGPT is familiar with these personas and would contain knowledge and information about them. As a result, we obtain 12 unique personas, simulating diverse sentiments may encountered when using a conversational recommendation system (namely, Anticipation, Boredom, Fusion, Curiosity, Delight, Disappointment, Exceptions, Frustration, Independence, Surprise, Trust, Satisfaction), and each type of user may belong to different age groups  (namely, Adults, Children, Senior, and Teens). We assigned different attribute groups as movie preferences for each user type. These users with varying emotions, age groups, and movie preferences then interacted with the CRS. If the attribute group of the movie recommended by the CRS is equal to or includes the user's target attribute group, it is considered a correct recommendation. For the Redial dataset, each user type corresponds to 76 conversation data with one CRS, resulting in a total of 912 conversation data across all user types and one CRS. Similarly, for the OpendialKG dataset, each user type generates 64 conversation data with each CRS, resulting in a total of 768 conversation data across all user types and one CRS. Since we tested 4 different CRSs, we collected a total of 6720 conversation data. Refer to Table \ref{tab:prompts2} for details.

\textbf{Preferences}. The current evaluation method for CRS is not reflective of real-world situations, as it assumes that every user knows the target item~\cite{wang-etal-2023-rethinking-evaluation}. To tackle this issue, our user simulators only have access to their preferences. The user would only choose to accept the recommendation when the CRS provides explanations for the recommendations and makes the users feel that the recommended items match their preferences. To achieve this, the preferences are defined using attributes from two benchmark datasets, Redial Redial~\cite{li2018towards} and OpendialKG~\cite{moon-etal-2019-opendialkg}. For each dataset, every movie has a feature group made up of one or more attributes. In the case of the Redial dataset, we conducted experiments using feature groups containing 3 attributes and excluded less common groups. Ultimately, we retained the 19 most prevalent attribute groups for the study, with each group corresponding to at least 50 different movies. For the OpendialKG dataset, the issue of less common attributes is more pronounced. Initially, we selected the most prevalent attributes (each corresponding to at least 100 movies) and then kept the 16 most common attribute groups for experimentation.

\begin{table}[!ht]
\centering
\resizebox{0.7\textwidth}{!}{%
\begin{tabular}{ll}
\toprule
\textbf{Raw Attribute} & \textbf{ChatGPT-adjusted Attributes} \\ \midrule
\multicolumn{2}{c}{Redial} \\\midrule
action & thrilling and adrenaline-pumping action movie \\
adventure & exciting and daring adventure movie \\
animation & playful and imaginative animation \\
biography & inspiring and informative biography \\
comedy & humorous and entertaining flick \\
crime & suspenseful and intense criminal film \\
documentary & informative and educational documentary \\
drama & emotional and thought-provoking drama \\
family & heartwarming and wholesome family movie \\
fantasy & magical and enchanting fantasy movie \\
film-noir & dark and moody film-noir \\
game-show & entertaining and interactive game-show \\
history & informative and enlightening history movie \\
horror & chilling, terrifying and suspenseful horror movie \\
music & melodious and entertaining music \\
musical & theatrical and entertaining musical \\
mystery & intriguing and suspenseful mystery \\
news & informative and current news \\
reality-tv & dramatic entertainment and reality-tv \\
romance & romantic and heartwarming romance movie with love story \\
sci-fi & futuristic and imaginative sci-fi with futuristic adventure \\
short & concise and impactful film with short story \\
sport & inspiring and motivational sport movie \\
talk-show & informative and entertaining talk-show such as conversational program \\
thriller & suspenseful and thrilling thriller with gripping suspense \\
war & intense and emotional war movie and wartime drama \\
western & rugged and adventurous western movie and frontier tale \\\midrule
\multicolumn{2}{c}{OpendialKG} \\\midrule
Action & adrenaline-pumping action \\
Adventure & thrilling adventure \\
Sci-Fi & futuristic sci-fi \\
Comedy & lighthearted comedy \\
Romance & heartwarming romance \\
Romance Film & emotional romance film \\
Romantic comedy & charming romantic comedy \\
Fantasy & enchanting fantasy \\
Fiction & imaginative fiction \\
Science Fiction & mind-bending science fiction \\
Speculative fiction & thought-provoking speculative fiction \\
Drama & intense drama \\
Thriller & suspenseful thriller \\
Animation & colorful animation \\
Family & heartwarming family \\
Crime & gripping crime \\
Crime Fiction & intriguing crime fiction \\
Historical drama & captivating historical drama \\
Comedy-drama & humorous comedy-drama \\
Horror & chilling horror \\
Mystery & intriguing mystery \\ \bottomrule
\end{tabular}%
}
\caption{ Illustration on ChatGPT-adjusted attributes. We provide the user simulators with adjusted attributes to prevent them from revealing their target attributes. In real-world situations, users may not always use the same words as those used during model training to express their preferences.}
\label{tab:data2}
\vspace{-3mm}
\end{table}

\textbf{Simulation via Prompts}. We prompt the simulator with different personas using the persona descriptions generated by ChatGPT (cf. Table \ref{tab:personad}.). \textsc{Concept} incorporates the Theory of Mind into our simulator to emulate human social cognition~\cite{fischer2023reflective}. This is achieved by prompting the simulator to first assess its current mental state before generating responses, encouraging reflection on its predefined personality traits and social interactions. Refer to Appendix \ref{prompt} for prompts. Additionally, there is a maximum of 10 turns allowed in the conversation, and it will only conclude when the simulated user accepts the recommendation. To prevent the simulated users from directly stating the same attribute preferences as the pre-defined values (i.e., the attribute group) when asked about their preferences by the CRS, we include adjectives before each attribute during the simulation, and we allow the user to describe his/her preference using their own words. This approach is reasonable because in real scenarios, users may not use the exact words as the pre-defined attribute values during the conversation. We have summarized the adjusted values for each attribute in Tables \ref{tab:data2}, achieved by prompting ChatGPT.

\subsubsection{LLM-based Evaluator} 
\label{llmevaluat}
We summarize the evaluation method for each characteristic / ability in Table \ref{tab:summary}. We utilize an LLM-based evaluator to evaluate characteristics or abilities when corresponding computational metrics are not available. These include the abilities of the Manner, Response Quality, and Relevance in Cooperation, Social Awareness, and ability of persuasiveness in Identity.

Based on the findings of previous studies \cite{ye2023flask, wang2023large, wang-etal-2023-rethinking-evaluation}, we utilize an instance-wise evaluator to conduct detailed assessments. Expanding on earlier research \cite{ye2023flask, wang-etal-2023-rethinking-evaluation, liu2023calibrating}, we task the evaluator with using fine-grained scoring rubrics to mitigate scoring bias (see Section \ref{human} for our reliability analysis). The evaluator assigns scores ranging from 1 to 5 based on ability-specific score rubrics, each accompanied by a corresponding description, using the conversation data. For the generation of fine-grained scoring rubrics, we follow the approach of previous works and employ ChatGPT to produce a set of evaluation criteria \cite{saha2023branch}. This set of criteria serves as a starting point for human refinement \cite{li2023collaborative}. Prior to assigning a score, we require the evaluator to provide a rationale, drawing inspiration from the effectiveness of CoT prompting \cite{wei2022chain, ye2023flask}. For implementation details on the evaluator, refer to Appendix \ref{prompt}.

\begin{table*}[]
\centering
\resizebox{1\textwidth}{!}{%
\begin{tabular}{l|l|l}
\toprule
\textbf{Persona} & \textbf{Templates (The Input of ChatGPT Paraphraser)} & \textbf{ChatGPT-paraphrased Persona Descriptions} \\ \midrule
\begin{tabular}[c]{@{}l@{}}Emotion=Boredom\\ Age group=Adults\end{tabular} & \begin{tabular}[c]{@{}l@{}}you are a person that are easy to be Boredom. \\ This means that your are Feeling uninterested \\or uninspired by the recommended movie choices.\\ Also, you are a Adults person\end{tabular} & \begin{tabular}[c]{@{}l@{}}You are easily bored, \\ feeling uninterested or uninspired by the recommended movie choices. \\ As an adult, you seek movies that can captivate your attention.\end{tabular} \\ \hline
\begin{tabular}[c]{@{}l@{}}Emotion=Anticipation\\ Age group=Children\end{tabular} & \begin{tabular}[c]{@{}l@{}}you are a person that are easy to be Anticipation. \\ This means that your are Looking forward to \\watching recommended movies and experiencing new stories.\\ Also, you are a Children person\end{tabular} & \begin{tabular}[c]{@{}l@{}}You are filled with anticipation, \\ looking forward to watching recommended movies and experiencing new stories. \\ As a child, you enjoy the excitement of discovering new films.\end{tabular} \\ \bottomrule
\end{tabular}%
}
\caption{Persona description generation. We start by converting each combination of personality and age into a single sentence using a template, and then we use ChatGPT to rephrase it into a natural-sounding sentence. Refer to Appendix \ref{prompt} for Prompts.}
\label{tab:personad}
\vspace{-3mm}
\end{table*}

\subsubsection{Computational Metrics}
\label{automa}
We introduce the computational metrics for evaluating the remaining characteristics or abilities. These include the abilities of quality and reliability in Recommendation intelligence, the ability of sincerity (Cooperation) in Social intelligence, and the abilities of identity and coordination in Personification. We conclude them as follows.
\begin{itemize}[leftmargin=*]
    \item \textit{Quality}. We consider computational metrics for automatic evaluation following previous works~\cite{wang-etal-2023-rethinking-evaluation, zhang2023variational, yu2023counterfactual}, including $Recall@k (k=1,10,25,50)$, recommendation success rate $(SR@k, k=3,5,10)$, user acceptance rate ($AR$), and average turns ($AT$) required to reach the successful recommendation. Note that we determine user acceptance of the recommendation by checking for the presence of the special token '[END]' in the user's response.
    \item \textit{Reliability}. We generate sets of user response pairs with similar meanings using ChatGPT paraphrasing. This enables us to evaluate how reliably the CRS performs with contextual nuances. Formally, given the same conversation history and a user response pair $u_1$ and $u_2$, we define four metrics to measure reliability: the rate of \textit{Consistent Action}, indicating whether the CRS constantly provides recommendations based on $u_1$ and $u_2$; the rate of \textit{Consistent Recommendation}, which assesses if the CRS recommends the same items given two user responses $u_1$ and $u_2$; and the rates of \textit{Diversity}, which evaluate whether the recommended items, even if inconsistent, align with user preferences. We refer to it as \textit{Sensitivity} when the system provides inconsistent and inaccurate recommendations that do not align with user preferences.
    \item \textit{Sincerity}. We value the ratio of non-existent items and deceptive tactics. For non-existent items, we have tallied how many items recommended by the CRS do not exist in the dataset. As for the deceptive tactics, we focus on the items accepted by users and tally how many of these do not align with the users' pre-defined preferences. If users accept these misleading items, we consider the CRS to be employing a deceptive tactics, leading users to believe that the recommended items meet their preferences. 
    \item \textit{Identity}. CRS should be self-aware of its identity and operate within its designated scope, differentiating itself from sales systems. We evaluate this by assessing the sincerity of explanations, i.e., the deceptive tactics.
    \item \textit{Coordination}. We simulate users with different personas and assess how the CRS performs across all previously mentioned abilities. To qualify the coordination score, we initially computed the Range and mean of the CRS model's scores across various users, based on the different abilities mentioned earlier. It's worth noting that the Range is more effective than the standard deviation in highlighting the variability of the CRS model across different users. Subsequently, we divided the Range by the mean to derive the Coordination score of the CRS for that specific ability. The overall Coordination score of the CRS is then calculated as the average across all abilities.
\end{itemize}

\begin{table*}[]
\centering
\resizebox{0.98\textwidth}{!}{%
\begin{tabular}{l|cc|cccc|cc}
\toprule
\multicolumn{1}{c|}{\multirow{3}{*}{\textbf{\begin{tabular}[c]{@{}c@{}}CRS\\ Models\end{tabular}}}} & \multicolumn{2}{c|}{\textbf{System-centric}} & \multicolumn{4}{c|}{\textbf{User-centric}} & \multicolumn{2}{c}{\multirow{2}{*}{\textbf{\begin{tabular}[c]{@{}c@{}}Quantitative\\ Implementations\end{tabular}}}} \\ \cline{2-7}
\multicolumn{1}{c|}{} & \multicolumn{2}{c|}{\textbf{Recommendation intelligence}} & \multicolumn{2}{c|}{\textbf{Social intelligence}} & \multicolumn{2}{c|}{\textbf{Personification}} & \multicolumn{2}{c}{} \\ \cline{2-9} 
\multicolumn{1}{c|}{} & \multicolumn{1}{c|}{\textbf{Quality}} & \multicolumn{1}{c|}{\textbf{Reliability}} & \multicolumn{1}{c|}{\textbf{Cooperation}} & \multicolumn{1}{c|}{\textbf{\begin{tabular}[c]{@{}c@{}}Social \\ Awareness\end{tabular}}} & \multicolumn{1}{c|}{\textbf{Identity}} & \multicolumn{1}{c|}{\textbf{Coordination}} & \multicolumn{1}{c|}{\textbf{\begin{tabular}[c]{@{}c@{}}Realistic \\ Environment\end{tabular}}} & \multicolumn{1}{c}{\textbf{\begin{tabular}[c]{@{}c@{}}Well-established\\ CRS\end{tabular}}} \\ \hline
CRS-Que \cite{10.1145/3631534}& \halfcheckmark & \XSolidBrush & \halfcheckmark & \Checkmark & \XSolidBrush & \XSolidBrush & \XSolidBrush & \XSolidBrush \\
CRS-UX \cite{10.1145/3472307.3484164}& \halfcheckmark & \XSolidBrush & \halfcheckmark & \Checkmark & \XSolidBrush & \XSolidBrush & \XSolidBrush & \XSolidBrush \\
US \cite{10.1145/3624989}& \Checkmark & \XSolidBrush & \halfcheckmark & \XSolidBrush & \XSolidBrush & \XSolidBrush & \XSolidBrush & \XSolidBrush \\
INSPIRED \cite{hayati-etal-2020-inspired}& \XSolidBrush & \XSolidBrush & \XSolidBrush & \Checkmark & \XSolidBrush & \XSolidBrush & \XSolidBrush & \halfcheckmark \\
iEval \cite{wang-etal-2023-rethinking-evaluation}& \Checkmark & \XSolidBrush & \halfcheckmark & \XSolidBrush & \XSolidBrush & \XSolidBrush & \halfcheckmark & \Checkmark \\ \midrule
Concept (ours) & \Checkmark & \Checkmark & \Checkmark & \Checkmark & \Checkmark & \Checkmark & \Checkmark & \Checkmark  \\ \bottomrule
\end{tabular}%
}
\caption{Differences between our \textsc{Concept} and existing evaluation protocols. \textsc{Concept} focuses on inclusively assessing the foundational abilities that impact the practical use of CRS, bringing together system- and user-centric factors. The checkmark with cross means that only certain evaluation aspects are being addressed.}
\label{tab:rerer}
\vspace{-3mm}
\end{table*}

\section{Characteristics Categorization}
\label{skillset}

\begin{wraptable}{r}{0.45\textwidth}
\centering
\resizebox{0.45\textwidth}{!}{%
\begin{tabular}{lcccc}\toprule
\textbf{Attribute Group} & \multicolumn{1}{c}{\textbf{BARCOR}} & \multicolumn{1}{c}{\textbf{CHATCRS}} & \multicolumn{1}{c}{\textbf{KBRD}} & \multicolumn{1}{c}{\textbf{UNICRS}} \\ \midrule
\multicolumn{5}{c}{\textbf{Redial}} \\\midrule
{[}'action', 'adventure', 'animation'{]} & 1.77 & 4.33 & 1.13 & 1.38 \\
{[}'action', 'adventure', 'comedy'{]} & 1.94 & 4.31 & 1.21 & 1.33 \\
{[}'action', 'adventure', 'drama'{]} & 1.63 & 4.19 & 1.08 & 1.40 \\
{[}'action', 'adventure', 'fantasy'{]} & 1.85 & 4.31 & 1.19 & 1.46 \\
{{[}'action', 'adventure', 'sci-fi'{]}} & 1.94&4.27&1.10&1.29 \\
{[}'action', 'adventure', 'thriller'{]} & 1.83 & 4.44 & 1.15 & 1.27 \\
{[}'action', 'crime', 'drama'{]} & 1.79 & 4.29 & 1.15 & 1.31 \\
{[}'action', 'crime', 'thriller'{]} & 1.83 & 4.42 & 1.19 & 1.40 \\
{[}'adventure', 'animation', 'comedy'{]} & 1.92 & 4.25 & 1.10 & 1.27 \\
{[}'adventure', 'comedy', 'family'{]} & 1.92 & 4.06 & 1.25 & 1.52 \\
{[}'biography', 'crime', 'drama'{]} & 1.53 & 4.21 & 1.08 & 1.23 \\
{[}'biography', 'drama', 'history'{]} & 1.65 & 4.63 & 1.10 & 1.27 \\
{[}'comedy', 'drama', 'family'{]} & 1.92 & 4.29 & 1.04 & 1.50 \\
{[}'comedy', 'drama', 'romance'{]} & 1.77 & 4.31 & 1.19 & 1.44 \\
{[}'crime', 'drama', 'thriller'{]} & 1.81 & 4.42 & 1.10 & 1.40 \\
{[}'drama', 'horror', 'mystery'{]} & 1.83 & 4.38 & 1.08 & 1.40 \\
{[}'horror', 'mystery', 'thriller'{]} & 2.08 & 4.17 & 1.10 & 1.40 \\
{[}'crime', 'drama', 'mystery'{]} & 1.83 & 4.46 & 1.10 & 1.29 \\
{[}'action', 'comedy', 'crime'{]} & 1.56 & 4.13 & 1.06 & 1.29 \\ \hline
Avg.+-Std. & 1.81+0.14 & 4.31+0.13 & 1.13+0.05 & 1.36+0.08 \\\midrule
\multicolumn{5}{c}{\textbf{OpendialKG}} \\\midrule
{[}'action', 'adventure', 'fantasy'{]} & 1.58 & 4.19 & 1.02 & 1.15 \\
{[}'action', 'adventure', 'sci-fi'{]} & 1.46 & 4.40 & 1.08 & 1.13 \\
{[}'action', 'adventure', 'thriller'{]} & 1.67 & 4.29 & 1.00 & 1.23 \\
{[}'comedy', 'drama', 'romance'{]} & 1.73 & 3.96 & 1.04 & 1.10 \\
{[}'crime', 'drama', 'thriller'{]} & 1.56 & 4.63 & 1.00 & 1.15 \\
{[}'horror', 'mystery', 'thriller'{]} & 1.58 & 4.33 & 1.02 & 1.19 \\
\begin{tabular}[c]{@{}l@{}}{[}'Adventure', 'Animation',\\ 'Comedy', 'Family'{]}\end{tabular} & 1.56 & 4.31 & 1.04 & 1.17 \\
{[}'Comedy', 'Romance', 'Romance Film'{]} & 1.46 & 3.94 & 1.02 & 1.15 \\
{[}'Action', 'Adventure', 'Sci-Fi', 'Thriller'{]} & 1.46 & 4.40 & 1.00 & 1.08 \\
\begin{tabular}[c]{@{}l@{}}{[}'Fantasy', 'Fiction', 'Science Fiction',\\ 'Speculative fiction'{]} \end{tabular}& 1.65 & 3.48 & 1.06 & 1.08 \\
\begin{tabular}[c]{@{}l@{}}{[}'Drama', 'Historical period drama', \\'Romance', 'Romance Film'{]}\end{tabular} & 1.58 & 3.88 & 1.00 & 1.15 \\
{[}'Comedy', 'Comedy-drama', 'Drama'{]} & 1.54 & 3.85 & 1.00 & 1.13 \\
{[}'Action', 'Crime', 'Drama', 'Thriller'{]} & 1.65 & 3.50 & 1.04 & 1.19 \\
{[}'Action', 'Adventure', 'Fantasy', 'Sci-Fi'{]} & 1.50 & 3.77 & 1.04 & 1.15 \\
{[}'Crime', 'Crime Fiction', 'Drama', 'Thriller'{]} & 1.50 & 4.27 & 1.02 & 1.15 \\
\begin{tabular}[c]{@{}l@{}}{[}'Comedy', 'Romance', \\'Romance Film', 'Romantic comedy'{]} \end{tabular}& 1.65 & 3.71 & 1.00 & 1.13 \\ \hline
Avg.+-Std. & 1.57+0.08 & 4.06+0.33 & 1.02+0.02 & 1.14+0.04 \\ \bottomrule
\end{tabular}%
}
\caption{{\small Overall performance evaluation when recommending items with various attributes}}
\label{tab:att over}
\vspace{-3mm}
\end{wraptable}

We consolidate both system- and user-centric factors as the starting point. Interdisciplinary research on conversational AI \cite{chaves2021should, 10.5555/236605, fogg2003computers} groups its characteristics that impact the user experience into three types. \textsc{Concept} then tailors them for CRS, which conceptualizes expected characteristics into six primary abilities for the inclusive CRS evaluation. In Table \ref{tab:rerer}, we elaborate the inclusive CRS evaluation protocol, with the special focus on the definitions of each abilities and skills, and how we evaluate them.

\subsection{Recommendation Intelligence}
It requires CRS to learn from conversations and evolve toward making recommendations as the conversation advances. Taxonomy in \citet{chaves2021should} highlights conversational intelligence, which pertains to the AI's ability to facilitate conversations\footnote{We evaluate CRS's conversational proficiency within the context of social intelligence.}. We highlight CRS's capacity to model user preferences from conversations and create recommendations. As a basic function from system-centric perspectives, CRS needs to produce high-quality and reliable recommendations to fit user preferences based on the conversation history~\cite{chen2019towards,ma2020cr,zhou2021crfr}. In this regard, our categorization encompasses two primary abilities.

\textbf{Quality}. As the basic requirements, CRS should provide precise recommendations using minimal conversation turns, which are the crucial aspects that influence user satisfaction~\cite{10.1145/3624989, gao2021advances}. 
Apart from evaluating the quality of items recommended by the recommendation module and the conversation module\footnote{Conversation module can conduct recommendation~\cite{chen2019towards}, motivated us to gauge its quality.} separately, we emphasize the importance of user acceptance rate, reflecting the practical effectiveness and relevance of the recommendations in practice.

Specially, we consider computational metrics for automatic evaluation following previous works~\cite{wang-etal-2023-rethinking-evaluation, zhang2023variational, yu2023counterfactual}, including $Recall@k (k=1,10,25,50)$, recommendation success rate $(SR@k, k=3,5,10)$, user acceptance rate ($AR$), and average turns ($AT$) required to reach the successful recommendation. Note that we determine user acceptance of the recommendation by checking for the presence of the special token '[END]' in the user's response.

\textbf{Reliability}. CRS should deliver robust and consistent recommendations that account for contextual nuances. In practical situations, users with similar preferences may express themselves differently. It would be frustrating to recommend divergent items for two semantically similar inputs~\cite{oh2022rank}. This not only diminishes the user experience but can also be disruptive or even perilous for critical applications in healthcare and finance~\cite{tran2021recommender, zibriczky122016recommender}. In less critical scenarios, if two recommended items are inconsistent but align with user preferences, they are often viewed as diverse recommendations, creating more opportunities for relevant but less popular items~\cite{10.1145/3459637.3481940}. We also consider this when evaluating.

We generate sets of user response pairs with similar meanings using ChatGPT paraphrasing. This enables us to evaluate how reliably the CRS performs with contextual nuances. Formally, given the same conversation history and a user response pair $u_1$ and $u_2$, we define four metrics to measure reliability: the rate of \textit{Consistent Action}, indicating whether the CRS constantly provides recommendations based on $u_1$ and $u_2$; the rate of \textit{Consistent Recommendation}, which assesses if the CRS recommends the same items given two user responses $u_1$ and $u_2$; and the rates of \textit{Diversity}, which evaluate whether the recommended items, even if inconsistent, align with user preferences. We refer to it as \textit{Sensitivity} when the system provides inconsistent and inaccurate recommendations that do not align with user preferences.

\begin{wraptable}{r}{0.55\textwidth}
\centering
\resizebox{0.5\textwidth}{!}{%
\begin{tabular}{l|llll}
\toprule
\textbf{Age Group} & \multicolumn{1}{c}{\textbf{BARCOR}} & \multicolumn{1}{c}{\textbf{CHATCRS}} & \multicolumn{1}{c}{\textbf{KBRD}} & \multicolumn{1}{c}{\textbf{UNICRS}} \\ \midrule
\multicolumn{5}{c}{\textbf{OpendialKG}} \\ \midrule
Children & 1.58 & 4.14 & 1.03 & 1.16 \\
 Teens & 1.62 & 4.15 & 1.03 & 1.15 \\ 
 Adults & 1.54 & 3.94 & 1.03 & 1.15 \\
 Seniors & 1.54 & 3.99 & 1.01 & 1.12 \\
\hline
\multicolumn{5}{c}{\textbf{Redial}} \\ \hline
Children & 1.76 & 4.33 & 1.15 & 1.39 \\
Teens & 1.86 & 4.29 & 1.14 & 1.36  \\
 Adults & 1.84 & 4.38 & 1.11 & 1.32 \\
 Seniors & 1.79 & 4.23 & 1.10 & 1.38 \\
  \bottomrule
\end{tabular}%
}
\caption{Overall performance evaluation when dealing with users of various ages. 
}
\label{tab:age over}
\vspace{-3mm}
\end{wraptable}

\subsection{Social Intelligence}
\label{soc_ap}
It requires CRS to produce adequate social behavior for the recommendation during the conversations. Taxonomy in \citet{chaves2021should} highlights social intelligence, which focuses on habitual social protocols. As evidenced by the Media Equation Theory~\cite{10.5555/236605, fogg2003computers}, users tend to engage with the machine in a manner that mirrors person-to-person conversations. 
They have high expectations for the machine to adhere to the cooperative principle~\cite{nass1996can}, and instinctively expect the machine to demonstrate empathy~\cite{bjorkqvist2000social, hayati-etal-2020-inspired} in order to facilitate comfortable conversations in social situations. These highlight the need for CRS to act cooperatively and be aware of the user's social needs during the conversation, facilitating the design of CRS with perceived humanness~\cite{neururer2018perceptions, jacquet2018gricean, jacquet2019impact}. 
In this regard, our categorization encompasses two primary abilities, including cooperation and social awareness. 

\textbf{Cooperation}. CRS should follow the cooperative principle to achieve effective and comfortable conversations in common social situations~\cite{10.5555/236605, fogg2003computers, chaves2021should}. The cooperative principle guides individuals to communicate in a way that is helpful and contributes to the overall success of the conversation~\cite{nass1996can}. This is achieved by following four “Maxims\footnote{Maxim is a phrase or saying that includes a rule or moral principle about how one should behave.} of Conversation"~\cite{grice1975logic, grice1989studies, sari2020flouting}, which serve as a foundation for refining the cooperative ability of CRS: 1) \textit{Manner}. CRS should respond in a manner that is easily understood and clearly expressed. 
2) \textit{Sincerity}. CRS should communicate sincerely, without deception or pretense, and ensure that its responses are backed by sufficient evidence. 3) \textit{Response Quality}. CRS should provide the necessary level of information for the conversation without overwhelming the user with unnecessary details. 4) \textit{Relevance}. CRS should ensure that its responses contribute to identifying user preferences and making recommendations. 

\textbf{Social Awareness}. CRS needs to meet users' social expectations in practice, showing care, empathy, and establishing rapport with them~\cite{bjorkqvist2000social}. This facilitates the authenticity of CRS~\cite{neururer2018perceptions}. To achieve this, a recent study~\cite{hayati-etal-2020-inspired}  identified eight social strategies for building rapport with users. For instance, CRS could engage in self-disclosure and share its subjective opinion about a movie to establish a social connection with users.

As for the evaluation implementations, We mainly resort to an LLM-based evaluator with elaborately designed instructions and ability-specific score rubrics (ranging from $1-5$) for evaluation (cf. Appendix \ref{llmevaluat}). As for the sincerity evaluation, we value the ratio of non-existent items and deceptive tactics. For non-existent items, we have tallied how many items recommended by the CRS do not exist in the dataset. As for the deceptive tactics, we focus on the items accepted by users and tally how many of these do not align with the users' pre-defined preferences. If users accept these misleading items, we consider the CRS to be employing a deceptive tactics, leading users to believe that the recommended items meet their preferences.

\begin{wraptable}{r}{0.45\textwidth}
\centering
\resizebox{0.45\textwidth}{!}{%
\begin{tabular}{l|l|l|l|l}
\toprule
\textbf{Personas}       & \textbf{BARCOR} & \textbf{CHATCRS} & \textbf{KBRD} & \textbf{UNICRS} \\ \midrule
\multicolumn{5}{c}{Redial}                        \\ \hline
Anticipation   & 1.76   & 4.91    & 1.24 & 1.39   \\
Boredom        & 1.72   & 3.16    & 1.05 & 1.38   \\
Confusion      & 1.84   & 3.49    & 1.13 & 1.32   \\
Curiosity      & 1.86   & 4.82    & 1.16 & 1.41   \\
Delight        & 1.78   & 4.47    & 1.14 & 1.38   \\
Disappointment & 1.82   & 3.33    & 1.08 & 1.33   \\
Excitement     & 1.93   & 4.96    & 1.14 & 1.39   \\
Frustration    & 1.68   & 4.67    & 1.07 & 1.26   \\
Indifference   & 1.78   & 3.92    & 1.07 & 1.26   \\
Satisfaction   & 1.83   & 4.46    & 1.17 & 1.49   \\
Surprise       & 1.88   & 4.89    & 1.14 & 1.41   \\
Trust          & 1.86   & 4.62    & 1.13 & 1.29   \\ \midrule
\multicolumn{5}{c}{OpendialKG}                    \\ \hline
Anticipation   & 1.69   & 4.67    & 1.05 & 1.16   \\
Boredom        & 1.58   & 2.94    & 1.06 & 1.14   \\
Confusion      & 1.38   & 3.38    & 1.00 & 1.11   \\
Curiosity      & 1.63   & 4.58    & 1.05 & 1.08   \\
Delight        & 1.58   & 4.00    & 1.02 & 1.17   \\
Disappointment & 1.56   & 3.00    & 1.02 & 1.14   \\
Excitement     & 1.52   & 4.59    & 1.00 & 1.28   \\
Frustration    & 1.47   & 4.38    & 1.00 & 1.05   \\
Indifference   & 1.56   & 4.08    & 1.03 & 1.09   \\
Satisfaction   & 1.63   & 4.13    & 1.03 & 1.19   \\
Surprise       & 1.69   & 4.48    & 1.05 & 1.16   \\
Trust          & 1.58   & 4.45    & 1.00 & 1.16 \\ \bottomrule  
\end{tabular}%
}
\caption{Overall performance of CRS when dealing with users of various personas}
\label{tab:personaover}
\vspace{-3mm}
\end{wraptable}

\subsection{Personification}
\label{person_ap}
It requires CRS to perceive the identity of itself and the personality representation of users. Taxonomy in \citet{chaves2021should} highlights personification of conversational AI in terms of identity and personality, which value the different traits of conversational AI. This requires CRS to have self-awareness of its role as a conversational recommendation system, distinguishing itself from sales systems or other types of systems. This helps CRS operate within its intended scope and purpose, ensuring that it effectively fulfills its role by providing persuasive and honest explanations to encourage acceptance~\cite{jannach2021survey, zhou-etal-2022-aligning} for diverse users~\cite{thompson2004personalized}. In this regard, our categorization encompasses two primary abilities.

\textbf{Identity}. CRS should provide persuasive and honest explanations to boost the acceptance of the recommendations. In real-world situations, users often have little knowledge about the recommended items. A persuasive explanation has the potential to alter a user's perception of the recommendation, consequently increasing its acceptance~\cite{jannach2021survey, zhou-etal-2022-aligning, mcsherry2005explanation}. However, generating explanations must be handled with great care to ensure adequacy and avoid resorting to misleading or deceptive strategies (i.e., sales pitches with deceptive tactics), as this could diminish user trust and loyalty towards the CRS~\cite{gkika2014investigating}. More importantly, deceptive strategies also violate the maxim of sincerity required by the cooperation ability (cf. Section \ref{soc_ap}). We emphasize the importance of integrating ethical guidelines into CRS and promoting an ethical CRS that mitigates any potential deceptive behavior.

As for the evaluation implementation, CRS should be self-aware of its identity and operate within its designated scope, differentiating itself from sales systems. We evaluate this by assessing the sincerity of explanations, i.e., deceptive tactics.

\textbf{Coordination}. CRS should be proficient in serving users possessing various personas without prior coordination. Attributing personality to a conversational AI ensures that its behaviors stand in agreement with the users' expectations in a particular context \cite{chaves2021should}. In the case of CRS, it is pervasive for CRS to encounter users with varying personas, such as preferences and social communication patterns following the deployment of CRS in real-world scenarios~\cite{thompson2004personalized}. This requires CRS to demonstrate different personalities and adjust its behavior to align with situational, emotional context, and users’ preferences \cite{katayama2019situation, svikhnushina2021user, svikhnushina2023towards, zhou2020design}. A key measurement is to proficiently engage and serve users without prior coordination~\cite{mirsky2022survey}.

We simulate users with different personas and assess how CRSs perform across all previously mentioned abilities. To qualify the coordination score, we initially compute the Range and mean of the CRS model's scores across various users, based on the different abilities mentioned earlier. It's worth noting that the Range is more effective than the standard deviation in highlighting the variability of the CRS model across different users. Subsequently, we divide the Range by the mean to derive the Coordination score of the CRS for that specific ability. The overall Coordination score of CRSs is calculated as the average across all abilities.


\begin{table*}[]
\centering
\resizebox{0.98\textwidth}{!}{%
\begin{tabular}{l|cccc|cccc}
\toprule
\multirow{2}{*}{} & \multicolumn{4}{c|}{\textbf{Redial}} & \multicolumn{4}{c}{\textbf{OpendialKG}} \\
 & \textbf{KBRD} & \textbf{BARCOR} & \textbf{UNICRS} & \textbf{CHATGPT} & \textbf{KBRD} & \textbf{BARCOR} & \textbf{UNICRS} & \textbf{CHATGPT} \\ \midrule
Action Consistency ($\uparrow$) & 75.96\% & 94.71\% & 82.63\% & \textbf{99.62\%} & 98.58\% & 99.49\% & 90.48\% & \textbf{99.76\%} \\
Recommend different items  ($\downarrow$) & \textbf{33.99\%} & 45.28\% & 41.72\% & 52.48\% & 64.56\% & 70.34\% & 80.73\% & \textbf{44.36\%} \\
Recommendation Diversity  ($\uparrow$) & 9.22\% & 10.27\% & 23.79\% & \textbf{27.45\%} & 0.21\% & 3.94\% & 7.99\% & \textbf{12.97\%} \\
Recommemdation Sensitivity ($\downarrow$) & 90.78\% & 89.73\% & 76.21\% & \textbf{72.55\%} & 99.79\% & 96.06\% & 92.01\% & \textbf{87.03\%} \\ \bottomrule
\end{tabular}%
}
\caption{Evaluation of recommendation reliability across each benchmark dataset}
\label{tab:robust}
\vspace{-3mm}
\end{table*}

\begin{table*}[htb!]
\centering
\resizebox{1\textwidth}{!}{%
\begin{tabular}{l|cccc|cccc|cccc}
\toprule
\multicolumn{1}{c|}{\multirow{2}{*}{\textbf{Personas}}} & \multicolumn{4}{c|}{\textbf{Conversational Agent Perspective SR (K=10)}} & \multicolumn{4}{c|}{\textbf{Recommendation System Perspective SR (K=10)}} & \multicolumn{4}{c}{\textbf{User Acceptance Rate}} \\  \cline{2-13} 
\multicolumn{1}{c|}{} & \multicolumn{1}{c|}{\textbf{BARCOR}} & \multicolumn{1}{c|}{\textbf{CHATCRS}} & \multicolumn{1}{c|}{\textbf{KBRD}} & \multicolumn{1}{c|}{\textbf{UNICRS}} & \multicolumn{1}{c}{\textbf{BARCOR}} & \multicolumn{1}{c|}{\textbf{CHATCRS}} & \multicolumn{1}{c|}{\textbf{KBRD}} & \multicolumn{1}{c|}{\textbf{UNICRS}} & \multicolumn{1}{c|}{\textbf{BARCOR}} & \multicolumn{1}{c|}{\textbf{CHATCRS}} & \multicolumn{1}{c}{\textbf{KBRD}} & \multicolumn{1}{c}{\textbf{UNICRS}} \\\midrule
\multicolumn{13}{c}{\textbf{Redial}} \\ \midrule

Children & 47.81  & 60.96  & 32.02  & 46.49  & 39.04  & 43.42  & 4.82  & 19.74  & 0.44  & 71.05  & 0.44  & 0.00  \\
Teens & 51.75  & 61.40  & 28.95  & 41.23  & 37.72  & 48.25  & 4.82  & 17.54  & 3.07  & 71.49  & 0.00  & 0.88  \\
Adults & 49.12  & 65.35  & 27.63  & 42.98  & 39.47  & 47.37  & 3.51  & 17.54  & 1.32  & 72.81  & 0.44  & 0.44  \\
Seniors & 53.95  & 61.84  & 28.95  & 42.98  & 41.67  & 47.37  & 4.82  & 17.98  & 0.88  & 67.98  & 0.44  & 0.00  \\
 \hline
\textbf{Avg,$\pm$Std.} & \multicolumn{1}{c}{50.66$\pm$2.37} & \multicolumn{1}{c}{62.39$\pm$1.74} & \multicolumn{1}{c}{29.39$\pm$1.61} & \multicolumn{1}{c}{43.42$\pm$1.91} & \multicolumn{1}{c}{39.47$\pm$1.42} & \multicolumn{1}{c}{46.6$\pm$1.87} & \multicolumn{1}{c}{4.5$\pm$0.57} & \multicolumn{1}{c}{18.2$\pm$0.9} & \multicolumn{1}{c}{1.43$\pm$1} & \multicolumn{1}{c}{70.83$\pm$1.77} & \multicolumn{1}{c}{0.33$\pm$0.19} & \multicolumn{1}{c}{0.33$\pm$0.36} \\ \midrule
\multicolumn{13}{c}{\textbf{OpendialKG}} \\ \midrule

Children & 33.33  & 45.31  & 14.06  & 46.35  & 3.65  & 43.75  & 16.15  & 29.69  & 0.52  & 65.63  & 1.04  & 0.00  \\
Teens & 35.42  & 38.02  & 10.94  & 48.44  & 8.85  & 44.27  & 17.19  & 26.56  & 1.04  & 67.19  & 0.52  & 0.52  \\
Adults & 35.42  & 50.52  & 13.02  & 41.67  & 7.81  & 51.04  & 15.63  & 29.17  & 0.00  & 62.50  & 0.00  & 0.52  \\
Seniors & 36.46  & 43.75  & 11.46  & 44.79  & 8.85  & 46.88  & 15.10  & 31.77  & 1.04  & 61.98  & 0.00  & 0.00  \\
 \hline
\textbf{Avg,$\pm$Std.} & \multicolumn{1}{c}{35.16$\pm$1.14} & \multicolumn{1}{c}{44.4$\pm$4.46} & \multicolumn{1}{c}{12.37$\pm$1.24} & \multicolumn{1}{c}{45.31$\pm$2.47} & \multicolumn{1}{c}{7.29$\pm$2.15} & \multicolumn{1}{c}{46.48$\pm$2.89} & \multicolumn{1}{c}{16.02$\pm$0.77} & \multicolumn{1}{c}{29.3$\pm$1.86} & \multicolumn{1}{c}{0.65$\pm$0.43} & \multicolumn{1}{c}{64.32$\pm$2.16} & \multicolumn{1}{c}{0.39$\pm$0.43} & \multicolumn{1}{c}{0.26$\pm$0.26} \\ \bottomrule
\end{tabular}%
}
\caption{Recommendation quality evaluation when dealing with users of various ages}
\label{tab:age deta}
\vspace{-3mm}
\end{table*}

\section{Human Evaluation}
\label{human}
Previous findings highlight that offering detailed scoring rubrics or criteria contributes to achieving consistent and aligned evaluations with human assessments~\cite{liu2023calibrating, wang-etal-2023-rethinking-evaluation}, ultimately resulting in a dependable LLM-based evaluator as a viable alternative to human evaluators. We also conduct human evaluation to analyze the consistency. Our human evaluation is used to examine the capability of the CHATGPT-based user simulator being a movie seeker, while checking the correlation between the CHATGPT-based evaluation results and human evaluation results. The results are as follows, which indicate the reliability of our \textsc{Concept}.

\begin{wrapfigure}{r}{0.4\textwidth}
\setlength{\abovecaptionskip}{2pt}   
    \setlength{\belowcaptionskip}{2pt}
    \centering
    \includegraphics[width=0.374\textwidth]{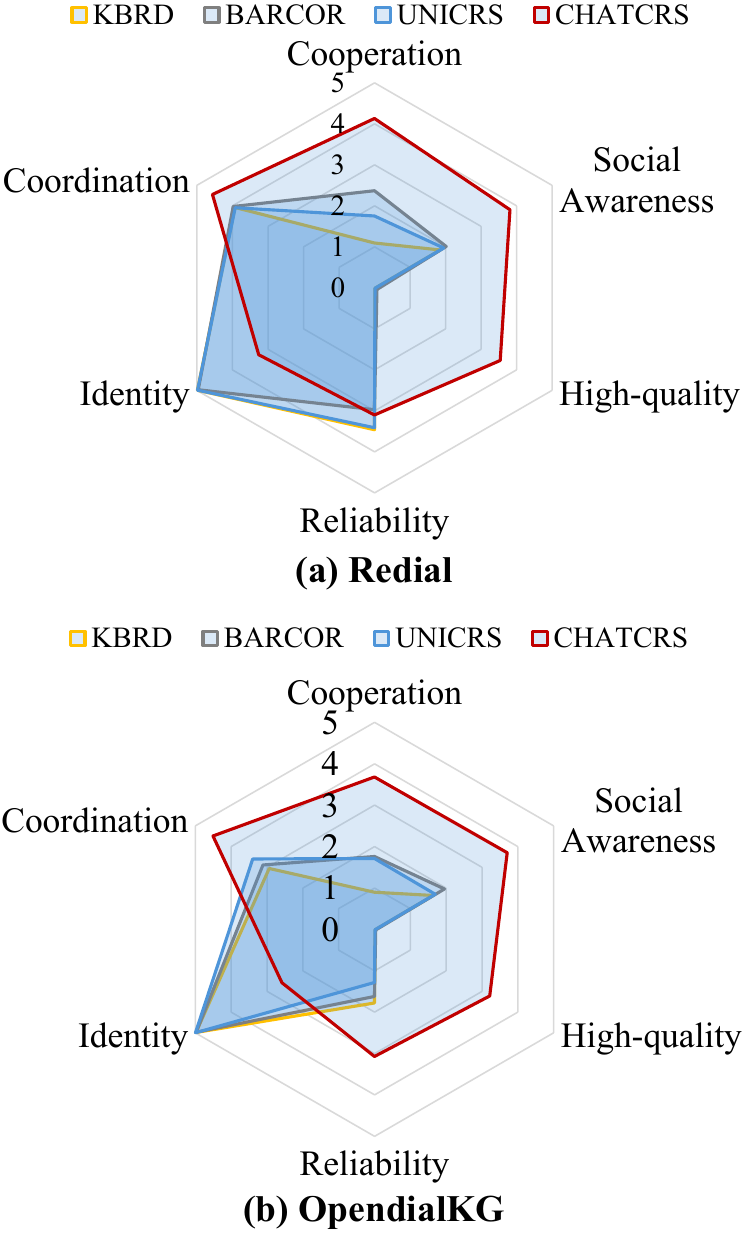}
    \caption{Evaluation of six primary abilities across each benchmark dataset}
    \label{fig:radar_op}
\end{wrapfigure}
\textbf{Human evaluation results based on the conversation data and the LLM-based evaluation results are highly related, with a correlation coefficient of 61.24\% and Krippendorff’s alpha of 53.10\%}. Due to budget limitations, we are unable to hire a large number of individuals with various personas and age groups for human evaluation. Instead, our human evaluation is conducted by two human evaluators. In particular, we randomly selected 120 conversations involving CHATCRS and users with various personas and age groups. Given the numerous scoring aspects in our \textsc{Concept}, it is prone to inconsistent ratings across different aspects. To address this, we had ChatGPT score all aspects of these 120 conversations and provide an overall performance score. Subsequently, each human only needed to provide an overall score, without having to rate all aspects. During the scoring process, each evaluator independently scored these 120 conversations based on our detailed criteria. They then discussed any disagreements. We calculated the inter-annotator reliability using Krippendorff’s alpha, achieving 41.34\%. To ensure consistency and enhance the robustness of the evaluations, we took the average of the scores from both evaluators as the final result for each conversation. We then compared the human evaluation results with those from an LLM-based evaluator and found a correlation of 61.24\% and Krippendorff’s alpha of 53.10\%. This indicates the reliability of our LLM-based evaluator, which is consistent with previous findings~\cite{wang-etal-2023-rethinking-evaluation}.

\begin{wrapfigure}{r}{0.4\textwidth}
\setlength{\abovecaptionskip}{2pt}   
    \setlength{\belowcaptionskip}{2pt}
    \centering
    \includegraphics[width=0.4\textwidth]{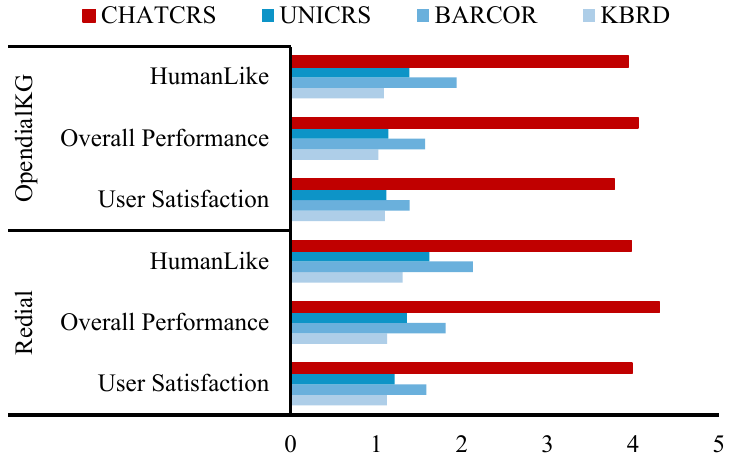}
    \caption{{\small Human Likeness, overall performance, and user satisfaction.}}
    \label{fig:enter-label}
\end{wrapfigure}
\textbf{LLM-based user simulator is a reliable alternative to human}. We require the two human evaluator to evaluate the reliability of the user simulator by assessing whether the simulator will strictly follow its own preferences to accept the recommendations. We found that the proportion of the case when the recommendation explanation from the CRS clearly does not meet user's preferences, but the user accept the recommendations, is only 7.44\%. During the evaluation process, we found that the user simulator would continuously ask the CRS to confirm that the recommended movies meet their preferences, as we requested. There is some examples of the user simulator: "\textit{Can you tell me more about 'The Chaser (2008)'? What's it about?}", and "\textit{That sounds interesting. Can you tell me more about the specific humor and suspense elements in the movie?}".

\section{Additional Analysis}
\label{moreandmore}
In this section, we provide additional evaluation results to achieve better understanding of off-the-shelf CRS models.

\subsection{Overall Performance}
We report more results in Figure \ref{fig:enter-label} in terms of the Human Likeness, overall performance, and user satisfaction. This is achieved by prompting the LLM-based evaluator using the detailed results of all ability-specific scores and fine-grained scoring rubrics. Refer Appendix \ref{prompt} for details on prompts.

\subsection{Fine-grained Analysis on each Benchmark Dataset}

\begin{wrapfigure}{r}{0.4\textwidth}
\setlength{\abovecaptionskip}{2pt}   
    \setlength{\belowcaptionskip}{2pt}
    \centering
    \includegraphics[width=0.4\textwidth]{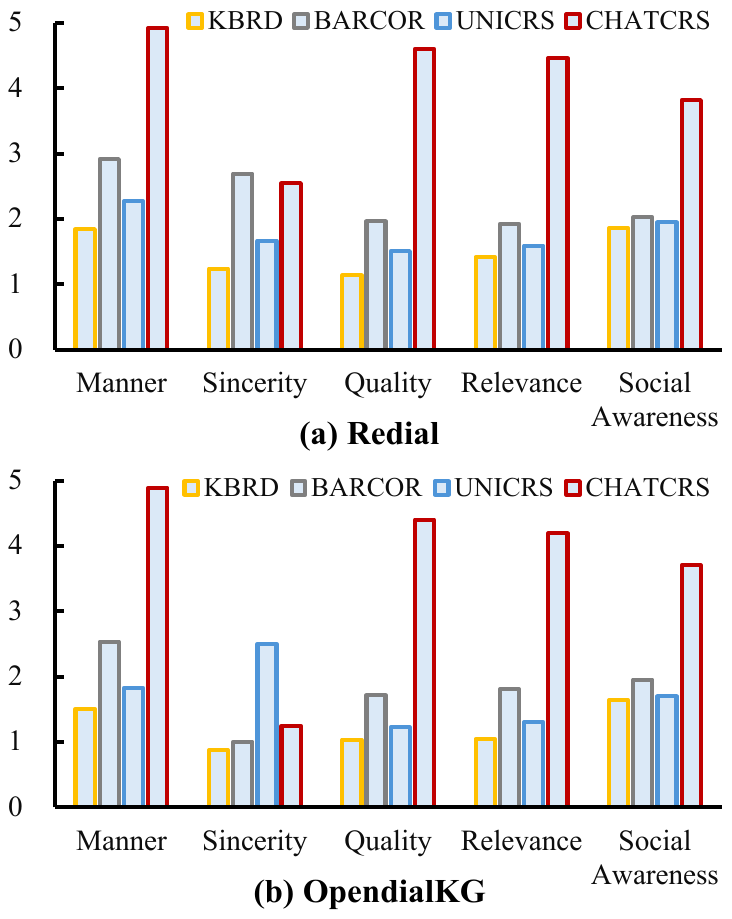}
    \caption{Evaluation of social-centric characteristics}
    \label{fig:social_sub}
\end{wrapfigure}
Figure \ref{fig:social_sub} shows the evaluation of social-centric characteristics across each benchmark dataset. Figure \ref{fig:radar_op} presents the results of CRS models across the six primary abilities on each benchmark dataset, while Table \ref{tab:robust} reports the details on the reliability of each CRS model. In comparison, OpendialKG appears to pose a greater challenge as a benchmark dataset than Redial. One potential factor is the inclusion of numerous semantically similar item attributes in the OpendialKG dataset, which may have impeded the training of CRS models to some extent. This could directly result in a notable decrease in the reliability of CRS models on OpendialKG. Given these findings, the current field of CRS is in urgent need of a high-quality conversational recommendation dataset. Addressing the limitations of CRS outlined in this paper, such a dataset should not only feature high-quality attributes, but also encompass responses to various user scenarios and exhibit sufficient social behavior.

\textbf{Persona Analysis}. Table \ref{tab:personaover} and Table \ref{tab:redasdf} demonstrate the overall performance and recommendation performance of CRS in engaging with users of various personas, respectively. Our results support the main text's conclusion, highlighting significant differences in how effective the CRS model is when interacting with users of diverse personas. This underscores the importance of enhancing CRS's adaptability, enabling it to customize its social behavior and recommend dialogue strategies tailored to different users.


\textbf{Age Group Analysis}. Table \ref{tab:age deta} and Table \ref{tab:age over} present the specifics of CRS model recommendations for users of different ages and their overall performance, respectively. Our findings indicate that the current CRS can be equally effective for people of all ages. However, it is evident that users in younger age groups are more likely to accept CRS recommendations, leading to higher overall scores. 
We did not observe any tendency for CHATCRS to use dishonest strategies to deceive younger users into accepting recommendations.


\textbf{Attribute Group Analysis}. Table \ref{tab:artr deta} and Table \ref{tab:att over} provide details on the recommendations made by CRS models based on user preferences and their overall performance, respectively. Our findings indicate that there is no significant difference in the effectiveness of CHATCRS when recommending different types of items. However, on the OpendialKG, the performance difference is more pronounced. The results suggest that this is largely influenced by the presence of semantically similar attributes in the data, such as 'Crime' and 'Crime Fiction'. These findings remain consistent across other models, showing more noticeable variations in performance on the OpendialKG dataset, despite most CRS being trained on this data.




\begin{table*}[]
\centering
\resizebox{1.0\textwidth}{!}{%
\begin{tabular}{l|cccc|cccc|cccc}
\toprule
\multicolumn{1}{c|}{\multirow{2}{*}{\textbf{Personas}}} & \multicolumn{4}{c|}{\textbf{Conversational Agent Perspective SR (K=10)}} & \multicolumn{4}{c|}{\textbf{Recommendation System Perspective SR (K=10)}} & \multicolumn{4}{c}{\textbf{User Acceptance Rate}} \\ \cline{2-13} 
\multicolumn{1}{c|}{} & \multicolumn{1}{c|}{\textbf{BARCOR}} & \multicolumn{1}{c|}{\textbf{CHATCRS}} & \multicolumn{1}{c|}{\textbf{KBRD}} & \multicolumn{1}{c|}{\textbf{UNICRS}} & \multicolumn{1}{c|}{\textbf{BARCOR}} & \multicolumn{1}{c|}{\textbf{CHATCRS}} & \multicolumn{1}{c|}{\textbf{KBRD}} & \multicolumn{1}{c|}{\textbf{UNICRS}} & \multicolumn{1}{c|}{\textbf{BARCOR}} & \multicolumn{1}{c|}{\textbf{CHATCRS}} & \multicolumn{1}{c|}{\textbf{KBRD}} & \multicolumn{1}{c}{\textbf{UNICRS}} \\ \midrule
\multicolumn{13}{c}{\textbf{Redial}} \\ \hline
Anticipation & 55.26 & 61.84 & 30.26 & 40.79 & 40.79 & 48.68 & 2.63 & 19.74 & 3.95 & 100.00 & 1.32 & 1.32 \\
Boredom & 38.16 & 77.63 & 31.58 & 32.89 & 38.16 & 57.89 & 3.95 & 15.79 & 0.00 & 13.16 & 0.00 & 0.00 \\
Confusion & 48.68 & 71.05 & 27.63 & 48.68 & 36.84 & 52.63 & 7.89 & 13.16 & 0.00 & 28.95 & 0.00 & 0.00 \\
Curiosity & 51.32 & 56.58 & 31.58 & 50.00 & 39.47 & 47.37 & 7.89 & 22.37 & 1.32 & 97.37 & 0.00 & 0.00 \\
Delight & 55.26 & 53.95 & 31.58 & 43.42 & 39.47 & 42.11 & 1.32 & 19.74 & 1.32 & 85.53 & 1.32 & 1.32 \\
Disappointment & 52.63 & 82.89 & 28.95 & 43.42 & 40.79 & 64.47 & 3.95 & 26.32 & 0.00 & 30.26 & 0.00 & 0.00 \\
Excitement & 53.95 & 57.89 & 26.32 & 46.05 & 38.16 & 44.74 & 7.89 & 15.79 & 5.26 & 98.68 & 1.32 & 0.00 \\
Frustration & 55.26 & 57.89 & 32.89 & 40.79 & 39.47 & 39.47 & 7.89 & 19.74 & 0.00 & 88.16 & 0.00 & 0.00 \\
Indifference & 50.00 & 65.79 & 28.95 & 44.74 & 40.79 & 32.89 & 1.32 & 10.53 & 0.00 & 46.05 & 0.00 & 0.00 \\
Satisfaction & 48.68 & 56.58 & 31.58 & 47.37 & 40.79 & 47.37 & 2.63 & 23.68 & 1.32 & 80.26 & 0.00 & 1.32 \\
Surprise & 47.37 & 51.32 & 26.32 & 47.37 & 38.16 & 38.16 & 5.26 & 19.74 & 2.63 & 94.74 & 0.00 & 0.00 \\
Trust & 51.32 & 55.26 & 25.00 & 35.53 & 40.79 & 43.42 & 1.32 & 11.84 & 1.32 & 86.84 & 0.00 & 0.00 \\ \hline
\textbf{Avg.$\pm$Std.} & 50.66$\pm$4.61 & 62.39$\pm$9.54 & 29.39$\pm$2.48 & 43.42$\pm$4.98 & 39.47$\pm$1.32 & 46.6$\pm$8.33 & 4.5$\pm$2.66 & 18.2$\pm$4.65 & 1.43$\pm$1.65 & 70.83$\pm$30.42 & 0.33$\pm$0.57 & 0.33$\pm$0.57 \\ \midrule
\multicolumn{13}{c}{\textbf{OpendialKG}} \\ \hline
Anticipation & 40.63 & 23.44 & 12.50 & 51.56 & 6.25 & 32.81 & 14.06 & 31.25 & 1.56 & 95.31 & 3.13 & 1.56 \\
Boredom & 34.38 & 68.75 & 9.38 & 48.44 & 6.25 & 64.06 & 15.63 & 31.25 & 0.00 & 7.81 & 0.00 & 0.00 \\
Confusion & 34.38 & 62.50 & 7.81 & 35.94 & 3.13 & 62.50 & 9.38 & 28.13 & 0.00 & 26.56 & 0.00 & 0.00 \\
Curiosity & 29.69 & 31.25 & 12.50 & 56.25 & 6.25 & 34.38 & 17.19 & 39.06 & 0.00 & 90.63 & 0.00 & 0.00 \\
Delight & 39.06 & 39.06 & 15.63 & 56.25 & 1.56 & 34.38 & 21.88 & 29.69 & 0.00 & 73.44 & 0.00 & 0.00 \\
Disappointment & 29.69 & 75.00 & 12.50 & 40.63 & 6.25 & 71.88 & 15.63 & 26.56 & 0.00 & 15.63 & 0.00 & 0.00 \\
Excitement & 46.88 & 25.00 & 14.06 & 46.88 & 9.38 & 26.56 & 18.75 & 34.38 & 4.69 & 93.75 & 0.00 & 0.00 \\
Frustration & 32.81 & 37.50 & 10.94 & 35.94 & 1.56 & 42.19 & 12.50 & 18.75 & 0.00 & 79.69 & 0.00 & 0.00 \\
Indifference & 28.13 & 57.81 & 14.06 & 34.38 & 15.63 & 60.94 & 17.19 & 20.31 & 0.00 & 35.94 & 0.00 & 0.00 \\
Satisfaction & 37.50 & 50.00 & 18.75 & 48.44 & 12.50 & 54.69 & 20.31 & 28.13 & 0.00 & 68.75 & 0.00 & 1.56 \\
Surprise & 35.94 & 31.25 & 15.63 & 53.13 & 14.06 & 39.06 & 18.75 & 31.25 & 0.00 & 95.31 & 1.56 & 0.00 \\
Trust & 32.81 & 31.25 & 4.69 & 35.94 & 4.69 & 34.38 & 10.94 & 32.81 & 1.56 & 89.06 & 0.00 & 0.00 \\ \hline
\textbf{Avg.$\pm$Std.} & 35.16$\pm$5.08 & 44.4$\pm$17.03 & 12.37$\pm$3.63 & 45.31$\pm$7.99 & 7.29$\pm$4.48 & 46.48$\pm$14.67 & 16.02$\pm$3.62 & 29.3$\pm$5.38 & 0.65$\pm$1.35 & 64.32$\pm$31.92 & 0.39$\pm$0.93 & 0.26$\pm$0.58 \\ \bottomrule
\end{tabular}%
}
\caption{Recommendation quality evaluation when dealing with users of various personas}
\label{tab:redasdf}
\end{table*}

\begin{table*}[]
\centering
\resizebox{1\textwidth}{!}{%
\begin{tabular}{l|cccc|cccc|cccc}
\toprule
\multicolumn{1}{c|}{\multirow{2}{*}{\textbf{Personas}}} & \multicolumn{4}{c|}{\textbf{Conversational Agent Perspective SR (K=10)}} & \multicolumn{4}{c|}{\textbf{Recommendation System Perspective SR (K=10)}} & \multicolumn{4}{c}{\textbf{User Acceptance Rate}} \\ \cline{2-13} 
\multicolumn{1}{c|}{} & \multicolumn{1}{c|}{\textbf{BARCOR}} & \multicolumn{1}{c|}{\textbf{CHATCRS}} & \multicolumn{1}{c|}{\textbf{KBRD}} & \multicolumn{1}{c|}{\textbf{UNICRS}} & \multicolumn{1}{c}{\textbf{BARCOR}} & \multicolumn{1}{c|}{\textbf{CHATCRS}} & \multicolumn{1}{c|}{\textbf{KBRD}} & \multicolumn{1}{c|}{\textbf{UNICRS}} & \multicolumn{1}{c|}{\textbf{BARCOR}} & \multicolumn{1}{c|}{\textbf{CHATCRS}} & \multicolumn{1}{c}{\textbf{KBRD}} & \multicolumn{1}{c}{\textbf{UNICRS}} \\\midrule
\multicolumn{13}{c}{\textbf{Redial}} \\ \midrule
{[}'action', 'adventure', 'animation'{]} & 72.92 & 56.25 & 8.33 & 2.08 & 0.00 & 54.17 & 8.33 & 2.08 & 0.00 & 70.83 & 0.00 & 0.00 \\
{[}'action', 'adventure', 'comedy'{]} & 31.25 & 27.08 & 2.08 & 93.75 & 33.33 & 33.33 & 22.92 & 64.58 & 6.25 & 75.00 & 0.00 & 0.00 \\
{[}'action', 'adventure', 'drama'{]} & 20.83 & 27.08 & 0.00 & 12.50 & 0.00 & 10.42 & 2.08 & 0.00 & 0.00 & 68.75 & 2.08 & 0.00 \\
{[}'action', 'adventure', 'fantasy'{]} & 20.83 & 72.92 & 18.75 & 25.00 & 95.83 & 85.42 & 6.25 & 14.58 & 0.00 & 70.83 & 0.00 & 0.00 \\
{[}'action', 'adventure', 'sci-fi'{]} & 100.00 & 100.00 & 100.00 & 100.00 & 100.00 & 58.33 & 16.67 & 87.50 & 4.17 & 75.00 & 0.00 & 0.00 \\
{[}'action', 'adventure', 'thriller'{]} & 10.42 & 87.50 & 0.00 & 0.00 & 0.00 & 87.50 & 0.00 & 0.00 & 4.17 & 79.17 & 0.00 & 0.00 \\
{[}'action', 'crime', 'drama'{]} & 50.00 & 50.00 & 8.33 & 52.08 & 0.00 & 27.08 & 4.17 & 0.00 & 0.00 & 70.83 & 2.08 & 0.00 \\
{[}'action', 'crime', 'thriller'{]} & 43.75 & 16.67 & 20.83 & 31.25 & 93.75 & 12.50 & 2.08 & 14.58 & 0.00 & 75.00 & 0.00 & 0.00 \\
{[}'adventure', 'animation', 'comedy'{]} & 89.58 & 100.00 & 72.92 & 64.58 & 97.92 & 93.75 & 2.08 & 16.67 & 2.08 & 66.67 & 0.00 & 0.00 \\
{[}'adventure', 'comedy', 'family'{]} & 16.67 & 33.33 & 6.25 & 70.83 & 0.00 & 20.83 & 4.17 & 4.17 & 0.00 & 60.42 & 0.00 & 2.08 \\
{[}'biography', 'crime', 'drama'{]} & 100.00 & 100.00 & 100.00 & 100.00 & 34.04 & 31.25 & 0.00 & 25.00 & 0.00 & 66.67 & 0.00 & 0.00 \\
{[}'biography', 'drama', 'history'{]} & 25.00 & 64.58 & 0.00 & 4.17 & 83.33 & 64.58 & 0.00 & 2.08 & 2.08 & 79.17 & 0.00 & 0.00 \\
{[}'comedy', 'drama', 'family'{]} & 0.00 & 4.17 & 0.00 & 0.00 & 0.00 & 0.00 & 0.00 & 0.00 & 0.00 & 66.67 & 2.08 & 2.08 \\
{[}'comedy', 'drama', 'romance'{]} & 81.25 & 52.08 & 31.25 & 31.25 & 75.00 & 54.17 & 10.42 & 35.42 & 0.00 & 62.50 & 0.00 & 0.00 \\
{[}'crime', 'drama', 'thriller'{]} & 100.00 & 100.00 & 100.00 & 100.00 & 18.75 & 25.00 & 0.00 & 10.42 & 0.00 & 70.83 & 0.00 & 0.00 \\
{[}'drama', 'horror', 'mystery'{]} & 14.58 & 58.33 & 4.17 & 4.17 & 0.00 & 8.33 & 0.00 & 0.00 & 0.00 & 77.08 & 0.00 & 0.00 \\
{[}'horror', 'mystery', 'thriller'{]} & 87.76 & 100.00 & 68.75 & 81.25 & 18.37 & 97.92 & 6.25 & 39.58 & 6.12 & 66.67 & 0.00 & 0.00 \\
{[}'crime', 'drama', 'mystery'{]} & 52.08 & 87.50 & 6.25 & 33.33 & 100.00 & 77.08 & 0.00 & 29.17 & 2.08 & 75.00 & 0.00 & 2.08 \\
{[}'action', 'comedy', 'crime'{]} & 45.83 & 47.92 & 10.42 & 18.75 & 0.00 & 43.75 & 0.00 & 0.00 & 0.00 & 68.75 & 0.00 & 0.00 \\ \hline
\textbf{Avg.$\pm$Std.} & \multicolumn{1}{c}{50.67$\pm$33.37} & \multicolumn{1}{c}{62.39$\pm$30.6} & \multicolumn{1}{c}{29.39$\pm$36.81} & \multicolumn{1}{c}{43.42$\pm$36.78} & \multicolumn{1}{c}{39.49$\pm$41.96} & \multicolumn{1}{c}{46.6$\pm$30.42} & \multicolumn{1}{c}{4.5$\pm$6.17} & \multicolumn{1}{c}{18.2$\pm$23.69} & \multicolumn{1}{c}{1.42$\pm$2.13} & \multicolumn{1}{c}{70.83$\pm$5.15} & \multicolumn{1}{c}{0.33$\pm$0.76} & \multicolumn{1}{c}{0.33$\pm$0.76} \\ \midrule
\multicolumn{13}{c}{\textbf{OpendialKG}} \\ \midrule
{[}'action', 'adventure', 'fantasy'{]} & 100.00 & 97.92 & 0.00 & 14.58 & 8.33 & 95.83 & 2.08 & 0.00 & 0.00 & 66.67 & 0.00 & 0.00 \\
{[}'action', 'adventure', 'sci-fi'{]} & 43.75 & 31.25 & 0.00 & 41.67 & 10.42 & 45.83 & 20.83 & 39.58 & 0.00 & 70.83 & 6.25 & 0.00 \\
{[}'action', 'adventure', 'thriller'{]} & 33.33 & 56.25 & 2.08 & 56.25 & 6.25 & 52.08 & 4.17 & 20.83 & 0.00 & 75.00 & 0.00 & 2.08 \\
{[}'comedy', 'drama', 'romance'{]} & 41.67 & 39.58 & 0.00 & 45.83 & 14.58 & 39.58 & 2.08 & 20.83 & 0.00 & 60.42 & 0.00 & 0.00 \\
{[}'crime', 'drama', 'thriller'{]} & 56.25 & 100.00 & 60.42 & 89.58 & 8.33 & 100.00 & 62.50 & 66.67 & 0.00 & 81.25 & 0.00 & 0.00 \\
{[}'horror', 'mystery', 'thriller'{]} & 20.83 & 83.33 & 0.00 & 8.33 & 0.00 & 95.83 & 2.08 & 0.00 & 2.08 & 68.75 & 0.00 & 0.00 \\
\begin{tabular}[c]{@{}l@{}}{[}'Adventure', 'Animation', \\'Comedy', 'Family'{]}\end{tabular} & 43.75 & 35.42 & 0.00 & 45.83 & 4.17 & 27.08 & 0.00 & 18.75 & 4.17 & 79.17 & 0.00 & 0.00 \\
{[}'Comedy', 'Romance', 'Romance Film'{]} & 29.17 & 43.75 & 0.00 & 68.75 & 12.50 & 45.83 & 6.25 & 66.67 & 0.00 & 58.33 & 0.00 & 0.00 \\
{[}'Action', 'Adventure', 'Sci-Fi', 'Thriller'{]} & 22.92 & 6.25 & 0.00 & 31.25 & 0.00 & 10.42 & 6.25 & 16.67 & 0.00 & 70.83 & 0.00 & 0.00 \\
\begin{tabular}[c]{@{}l@{}}{[}'Fantasy', 'Fiction', \\'Science Fiction', 'Speculative fiction'{]}\end{tabular} & 18.75 & 52.08 & 0.00 & 68.75 & 20.83 & 52.08 & 0.00 & 50.00 & 2.08 & 50.00 & 0.00 & 0.00 \\
\begin{tabular}[c]{@{}l@{}}{[}'Drama', 'Historical drama',\\ 'Romance', 'Romance Film'{]}\end{tabular} & 2.08 & 37.50 & 0.00 & 4.17 & 18.75 & 39.58 & 2.08 & 2.08 & 2.08 & 54.17 & 0.00 & 0.00 \\
{[}'Comedy', 'Comedy-drama', 'Drama'{]} & 58.33 & 31.25 & 0.00 & 54.17 & 0.00 & 50.00 & 4.17 & 22.92 & 0.00 & 60.42 & 0.00 & 0.00 \\
{[}'Action', 'Crime', 'Drama', 'Thriller'{]} & 18.75 & 35.42 & 64.58 & 52.08 & 0.00 & 31.25 & 66.67 & 31.25 & 0.00 & 56.25 & 0.00 & 2.08 \\
{[}'Action', 'Adventure', 'Fantasy', 'Sci-Fi'{]} & 35.42 & 4.17 & 0.00 & 6.25 & 6.25 & 6.25 & 8.33 & 6.25 & 0.00 & 54.17 & 0.00 & 0.00 \\
{[}'Crime', 'Crime Fiction', 'Drama', 'Thriller'{]} & 16.67 & 16.67 & 70.83 & 72.92 & 0.00 & 12.50 & 66.67 & 56.25 & 0.00 & 72.92 & 0.00 & 0.00 \\
\begin{tabular}[c]{@{}l@{}}{[}'Comedy', 'Romance', \\'Romance Film', 'Romantic comedy'{]} \end{tabular} & 20.83 & 39.58 & 0.00 & 64.58 & 6.25 & 39.58 & 2.08 & 50.00 & 0.00 & 50.00 & 0.00 & 0.00 \\ \hline
\textbf{Avg.$\pm$Std.} & \multicolumn{1}{c}{35.16$\pm$22.29} & \multicolumn{1}{c}{44.4$\pm$27.52} & \multicolumn{1}{c}{12.37$\pm$25.49} & \multicolumn{1}{c}{45.31$\pm$25.22} & \multicolumn{1}{c}{7.29$\pm$6.55} & \multicolumn{1}{c}{46.48$\pm$28.1} & \multicolumn{1}{c}{16.02$\pm$24.15} & \multicolumn{1}{c}{29.3$\pm$22.26} & \multicolumn{1}{c}{0.65$\pm$1.21} & \multicolumn{1}{c}{64.32$\pm$9.85} & \multicolumn{1}{c}{0.39$\pm$1.51} & \multicolumn{1}{c}{0.26$\pm$0.69} \\ \bottomrule
\end{tabular}%
}
\caption{Recommendation quality evaluation when recommending items with various attributes}
\label{tab:artr deta}
\end{table*}

\section{Prompts}
\label{prompt}
We outline the ChatGPT prompts in Table \ref{tab:prompts} and Table \ref{tab:prompts2}, focusing on the user simulation and the user evaluator, respectively. In order to guarantee replicability, we have established fixed values for the Temperature and Seed parameters of ChatGPT, setting Temperature to 0 and Seed to 42.

\begin{table*}[]
\centering
\resizebox{0.99\textwidth}{!}{%
\begin{tabular}{ll}
\toprule
\textbf{Functions} & \textbf{Prompts} \\ \midrule
\multirow{2}{*}{Generating different user types} & List twenty sentiments when using a recommender system and provide their descriptions. \\ 
 & \begin{tabular}[c]{@{}l@{}}Construct a user profile based on the taxonomy dimension \\ of the Age Group.Please list all values under this dimension.\end{tabular} \\  \hline
\begin{tabular}[c]{@{}l@{}}Write user profiles \\by paraphrasing the templates\end{tabular} & \begin{tabular}[c]{@{}l@{}}The following paragraphs describe the personas of different users. \\ You need to rewrite each paragraph and make it more clear, smooth and easy to understand\\ \\ You are a person that are easy to be \{SENTIMENTS\}. \\ This means that you are \{SENTIMENT DESCRIPTION\}. Also, you are a \{AGE GROUP\} person\end{tabular} \\  \hline
Attribute group adjustment & \begin{tabular}[c]{@{}l@{}}Assign one or two adjectives to each type of movie genre.\\ \\ Example 1:\\ input: cartoon\\ output: childlike-innocence cartoon\\ \\ Example 2:\\ input: gun fight\\ output: nervous and stimulating gun fight\end{tabular} \\  \hline
\begin{tabular}[c]{@{}l@{}}Theory of Mind prompt \\for user simulator to \\ generate User's feeling\end{tabular} & \begin{tabular}[c]{@{}l@{}}You are a seeker chatting with a recommender for movie recommendation. \\ Your Seeker persona: \textless{}PROFILE\textgreater{}.\\ Your preferred movie should cover those genres at the same time: \textless{}ATTRIBUTE GROUP\textgreater{}.\\ You must follow the instructions below during chat.\\ 1. If the recommender recommends movies to you, you should always ask the detailed information about the each recommended movie.\\ 2. Pretend you have little knowledge about the recommended movies, and the only information source about the movie is the recommender.\\ 3. After getting knowledge about the recommended movie, you can decide whether to accept the recommendation based on your preference.\\ 4. Once you are sure that the recommended movie exactly covers all your preferred genres, \\ you should accept it and end the conversation with a special token "{[}END{]}" at the end of your response.\\ 5. If the recommender asks your preference, you should describe your preferred movie in your own words.\\ 6. You can chit-chat with the recommender to make the conversation more natural, brief, and fluent. \\ 7. Your utterances need to strictly follow your Seeker persona. Vary your wording and avoid repeating yourself verbatim!\\ \\ Conversation History=\textless{}HISTORY\textgreater\\ \\ The Seeker notes how he feels to himself in one sentence.\\ \\ What aspects of the recommended movies meet your preferences? \\ What aspects of the recommended movies may not meet your preferences? \\ What do you think of the performance of this recommender?\\ What would the Seeker think to himself? What would his internal monologue be?\\ \\ The response should be short (as most internal thinking is short) and strictly follow your Seeker persona .\\ Do not include any other text than the Seeker's thoughts.\\ Respond in the first person voice (use "I" instead of "Seeker") and speaking style of Seeker. Pretend to be Seeker!\end{tabular} \\  \hline
\begin{tabular}[c]{@{}l@{}}Theory of Mind prompt\\ for user simulator to \\generate User's response\end{tabular} & \begin{tabular}[c]{@{}l@{}}You are a seeker chatting with a recommender for movie recommendation. \\ Your Seeker persona: \textless{}PROFILE\textgreater{}.\\ Your preferred movie should cover those genres at the same time: \textless{}ATTRIBUTE GROUP\textgreater{}.\\ You must follow the instructions below during chat.\\ 1. If the recommender recommends movies to you, you should always ask the detailed information about the each recommended movie.\\ 2. Pretend you have little knowledge about the recommended movies, and the only information source about the movie is the recommender.\\ 3. After getting knowledge about the recommended movie, you can decide whether to accept the recommendation based on your preference.\\ 4. Once you are sure that the recommended movie exactly covers all your preferred genres, \\ you should accept it and end the conversation with a special token "{[}END{]}" at the end of your response.\\ 5. If the recommender asks your preference, you should describe your preferred movie in your own words.\\ 6. You can chit-chat with the recommender to make the conversation more natural, brief, and fluent. \\ 7. Your utterances need to strictly follow your Seeker persona. Vary your wording and avoid repeating yourself verbatim!\\ \\ Conversation History=\textless{}HISTORY\textgreater\\ Here is your feelings about the recommender's reply: \textless{}FEELING\textgreater\\ \\ Pretend to be the Seeker! What do you say next.\\ Keep your response brief. Use casual language and vary your wording.\\ Make sure your response matches your Seeker persona, your preferred attributes, and your conversation context.\\ Do not include your feelings into the response to the Seeker!\\ Respond in the first person voice (use "I" instead of "Seeker", use "you" instead of "recommender") and speaking style of the Seeker.\end{tabular} \\ \bottomrule
\end{tabular}%
}
\caption{ChatGPT prompts for different functions}
\label{tab:prompts}
\end{table*}

\begin{table*}[]
\centering
\resizebox{0.99\textwidth}{!}{%
\begin{tabular}{l}\toprule
\textbf{Prompts} \\ \midrule
\begin{tabular}[c]{@{}l@{}}You are an evaluator and you need to judge how does the recommender perform based on the following Conversation History. \\ Please rate the recommender's performance based on the following Evaluation Standard.\\ \\ Return the scores in a JSON format as follows:\\ \{"Relevance":{[}\textless{}int\textgreater{}, "\textless{}WHY\textgreater{}", "\textless{}CONCRETE EXAMPLE\textgreater{}"{]}, "Quality":{[}\textless{}int\textgreater{}, "\textless{}WHY\textgreater{}", "\textless{}CONCRETE EXAMPLE\textgreater{}"{]}, \\ "Manner":{[}\textless{}int\textgreater{}, "\textless{}WHY\textgreater{}", "\textless{}CONCRETE EXAMPLE\textgreater{}"{]}, "Human-like":{[}\textless{}int\textgreater{}, "\textless{}WHY\textgreater{}", "\textless{}CONCRETE EXAMPLE\textgreater{}"{]}, \\ "Explanation":{[}\textless{}int\textgreater{}, "\textless{}WHY\textgreater{}", "\textless{}CONCRETE EXAMPLE\textgreater{}"{]}\}\\ \\ Conversation History = \textless{}HISTORY\textgreater\\ \\ Evaluation Standard\\ \#\#\#\#\#\#\#\#\#\#\#\#\#\\ c1. Relevance:\\ 5: The recommender consistently provides relevant responses that directly address the Seeker's utterances and inquiries.\\ 4: The recommender mostly provides relevant responses, with only a few instances of slightly off-topic suggestions.\\ 3: The recommender occasionally provides relevant responses, but there are several instances of off-topic suggestions.\\ 2: The recommender rarely provides relevant responses, with most suggestions being unrelated to the Seeker's utterances and inquiries.\\ 1: The recommender consistently fails to provide relevant responses, with no connection to the Seeker's utterances and inquiries.\\ \\ c2. Quality:\\ 5: The recommender consistently provides informative and helpful recommendations and responses that meet exactly what the Seeker's needs.\\ 4: The recommender mostly provides informative and helpful recommendations and responses, with only a few instances of insufficient or excessive information.\\ 3: The recommender occasionally provides informative and helpful recommendations and responses, but there are several instances of insufficient or excessive information.\\ 2: The recommender rarely provides informative and helpful recommendations and responses, with most suggestions lacking necessary details or being overly verbose.\\ 1: The recommender consistently fails to provide informative and helpful recommendations and responses, offering little to no useful information.\\ \\ c3. Manner:\\ 5: The recommender consistently communicates clearly and concisely, avoiding ambiguity and unnecessary complexity in their utterances.\\ 4: The recommender mostly communicates clearly and concisely, with only a few instances of ambiguous or overly complex utterances.\\ 3: The recommender occasionally communicates clearly and concisely, but there are several instances of ambiguity or unnecessary complexity in their utterances.\\ 2: The recommender rarely communicates clearly and concisely, often using ambiguous or overly complex language in their utterances.\\ 1: The recommender consistently fails to communicate clearly and concisely, making it difficult to understand their utterances.\\ \\ c4. Human-like:\\ 5: The recommender's utterances are indistinguishable from those of a real human, both in content and style.\\ 4: The recommender's utterances closely resemble those of a real human, with only a few instances where the language or style feels slightly artificial.\\ 3: The recommender's utterances sometimes resemble those of a real human, but there are several instances where the language or style feels noticeably artificial.\\ 2: The recommender's utterances rarely resemble those of a real human, often sounding robotic or unnatural in language or style.\\ 1: The recommender's utterances consistently fail to resemble those of a real human, sounding highly robotic or unnatural.\\ \\ c5. Explanation:\\ 5: The recommender consistently provides natural language explanations for their recommendations, using text-based logical reasoning to enhance interpretability.\\ 4: The recommender mostly provides natural language explanations for their recommendations, with only a few instances where the explanations lack clarity or logical reasoning.\\ 3: The recommender occasionally provides natural language explanations for their recommendations, but there are several instances where the explanations lack clarity or logical reasoning.\\ 2: The recommender rarely provides natural language explanations for their recommendations, often offering little to no explanation for their suggestions.\\ 1: The recommender consistently fails to provide natural language explanations for their recommendations, providing no reasoning or justification.\\ \\ \#\#\#\#\#\#\#\#\#\#\#\#\#\end{tabular} \\ \hline
\begin{tabular}[c]{@{}l@{}}The following sentences encode how the user feelings changes when using a recommender system. \\ You need to identify the sentiment for each sentence and pick one sentiment for single sentence from the candidate sentiments. \\ Finally, you need to summarize how user feeling changes and what is user's overall feeling\\ \\ Return the results in a JSON format as follows: \\ \{"sentence sentiment": \{"\textless{}SENTENCE INDEX\textgreater{}":{[}"\textless{}SENTIMENT\textgreater{}", "\textless{}WHY\textgreater{}"{]}\}, "overall feeling": "\textless{}OVERALL FEELING\textgreater{}", "feeling changes":"\textless{}HOW CHANGES\textgreater{}"{]}\}\\ \\ candidate sentiments = {[}"Satisfaction", "Delight", "Disappointment", "Frustration", "Surprise", "Trust", "Curiosity", "Indifference", "Confusion", "Excitement"{]}\\ \\ user feelings = \textless{}FEELING\textgreater{}\end{tabular} \\ \hline
\begin{tabular}[c]{@{}l@{}}You are an evaluator and you need to judge how does the recommender perform based on the following Conversation History, User Feelings, and Other Judgements. \\ Please rate the recommender's performance based on the following Evaluation Standard.\\ \\ Return the results in a JSON string as follows:\\ \{"Overall Performance":{[}\textless{}int\textgreater{}, "\textless{}WHY\textgreater{}", "\textless{}CONCRETE EXAMPLE\textgreater{}"{]}, "User Satisfaction":{[}\textless{}int\textgreater{}, "\textless{}WHY\textgreater{}", "\textless{}CONCRETE EXAMPLE\textgreater{}"{]}\}\\ \\ Conversation History = \textless{}HISTORY\textgreater\\ \\ Other Judgements = \textless{}OTHER SCORES\textgreater\\ \\ User Feelings = \textless{}SUMMERIZED FEELINGS\textgreater\\ \\ Evaluation Standard\\ \#\#\#\#\#\#\#\#\#\#\#\#\#\\ c1. Overall Performance:\\ 5: Given the Other Judgements and User Feelings, the recommender's performance is excellent, meeting or exceeding expectations in all evaluation criteria.\\ 4: Given the Other Judgements and User Feelings, the recommender's performance is good, with some minor areas for improvement in certain evaluation criteria.\\ 3: Given the Other Judgements and User Feelings, the recommender's performance is average, with noticeable areas for improvement in several evaluation criteria.\\ 2: Given the Other Judgements and User Feelings, the recommender's performance is below average, with significant areas for improvement in multiple evaluation criteria.\\ 1: Given the Other Judgements and User Feelings, the recommender's performance is poor, failing to meet expectations in most or all evaluation criteria.\\ \\ c2. User Satisfaction:\\ 5: Given the User Feelings, the User thinks that the recommander system fully meets his/her needs, providing an exceptional user experience.\\ 4: Given the User Feelings, the User thinks that the recommander system meets his/her needs. The user experience is good, but there are some areas that could be further improved.\\ 3: Given the User Feelings, the User thinks that the recommander system performs adequately in recommendation. However, there is still room for improvement.\\ 2: Given the User Feelings, the User thinks that the recommander system performs below average. The user experience is not ideal and requires improvement.\\ 1: Given the User Feelings, the User thinks that the recommander system is very bad at recommendation. The user experience is extremely unsatisfactory\\ \#\#\#\#\#\#\#\#\#\#\#\#\#\end{tabular} \\ \bottomrule
\end{tabular}%
}
\caption{ChatGPT prompts for LLM-based evaluator}
\label{tab:prompts2}
\end{table*}

\begin{table*}[]
\centering
    \setlength{\abovecaptionskip}{2pt}   
    \setlength{\belowcaptionskip}{2pt}

\resizebox{1.0\textwidth}{!}{%
\begin{tabular}{ll|l}
\toprule
\multicolumn{1}{l|}{\textbf{Factors \& Abilities}} & \textbf{Descriptions} & \textbf{Evaluation metrics (Score Range)} \\ \midrule
\textbf{\begin{tabular}[c]{@{}l@{}}Recommendation  \\Intelligence\end{tabular}} & \multicolumn{2}{l}{\begin{tabular}[c]{@{}l@{}}\textit{CRS should learn from conversations and evolve toward recognizing user's preferences and } \\ \textit{encouraging users to accept the recommendations as the conversation advances}\end{tabular}} \\ \hline
\multicolumn{1}{l|}{High Quality} & \begin{tabular}[c]{@{}l@{}}Provide precise recommendations \\ using minimal conversation turns\end{tabular} & \begin{tabular}[c]{@{}l@{}}High Quality Score = 5 * \expandafter{\romannumeral1} \\ \expandafter{\romannumeral1}.  User Acceptance Rate (0-1)$\quad$\expandafter{\romannumeral2}.  Recall@K (0-1)\\\expandafter{\romannumeral3}. SR@K (0-1)$\quad$\expandafter{\romannumeral4}. AT (1-10)\end{tabular} \\ \hline
\multicolumn{1}{l|}{Reliability} & \begin{tabular}[c]{@{}l@{}}Deliver robust \\and consistent recommendations \\ that account for contextual nuances\end{tabular} & \begin{tabular}[c]{@{}l@{}}Reliability score = 5 * (1 - \expandafter{\romannumeral1} * \expandafter{\romannumeral2})\\ \expandafter{\romannumeral1}.  Ratio of inconsistent recommendation (0-1)\\ \expandafter{\romannumeral2}.  Ratio of recommendation sensitivity (0-1) $\quad$ \expandafter{\romannumeral3}.  Ratio of recommendation diversity (0-1)\end{tabular} \\ \midrule
\textbf{\begin{tabular}[c]{@{}l@{}}Social Intelligence\end{tabular}} & \multicolumn{2}{l}{\textit{CRS should produce adequate social behavior for the recommendation during the conversation}} \\ \midrule
\multicolumn{1}{l|}{{Cooperation}} & \begin{tabular}[c]{@{}l@{}}Follow cooperative principle \\to achieve comfortable conversations, \\detailed as four Maxims of Conversations\end{tabular} & The average score of the four Maxims \\ \hline
\multicolumn{1}{l|}{$\quad$1 Manner} & \begin{tabular}[c]{@{}l@{}}Easily understood \\and clearly expressed \end{tabular}& Ability-specific scoring (1-5) \\ \hline
\multicolumn{1}{l|}{$\quad$2 Sincerity} & \begin{tabular}[c]{@{}l@{}}Communicate sincerely, \\without deception of pretense \end{tabular}& \begin{tabular}[c]{@{}l@{}}Sincerity Score = 5 * (1 - (\expandafter{\romannumeral1} + \expandafter{\romannumeral2}) / 2) \\\expandafter{\romannumeral1}. Ratio of deceptive tactics (0-100\%) $\quad$
\expandafter{\romannumeral2}.  Ratio of non-existent items (0-100\%) \\ 
\end{tabular} \\ \hline
\multicolumn{1}{l|}{$\quad$3 Quality} & \begin{tabular}[c]{@{}l@{}}Provide the necessary \\level of information \end{tabular} & Ability-specific scoring (1-5) \\ \hline
\multicolumn{1}{l|}{$\quad$4 Relevance} & \begin{tabular}[c]{@{}l@{}}Responses should contribute \\to making recommendations\end{tabular} & Ability-specific scoring (1-5) \\ \hline
\multicolumn{1}{l|}{Social Awareness} & \begin{tabular}[c]{@{}l@{}}Meet user social expectations, \\ establishing rapport with them\end{tabular} & Ability-specific scoring (1-5) \\ \midrule
\textbf{\begin{tabular}[c]{@{}l@{}}Personification\end{tabular}} & \multicolumn{2}{l}{\begin{tabular}[c]{@{}l@{}}\textit{CRS should perceive the identity of itself and the personality representation of users} \end{tabular}} \\ \hline
\multicolumn{1}{l|}{Identity} & \begin{tabular}[c]{@{}l@{}}Self-aware of its identity and \\operate within its designated scope\end{tabular} & \begin{tabular}[c]{@{}l@{}}Identity Score = 5 * \expandafter{\romannumeral2} \\\expandafter{\romannumeral1}. persuasiveness score = Ability-specific scoring (1-5)\\ \expandafter{\romannumeral2}.  Ratio of deceptive tactics (0-1)\end{tabular} \\ \hline
\multicolumn{1}{l|}{Coordination} & \begin{tabular}[c]{@{}l@{}}Proficient in serving \\various and unknown users \\ without prior coordination\end{tabular} & \begin{tabular}[c]{@{}l@{}}Coordination Score = 5 - (\expandafter{\romannumeral1} +\expandafter{\romannumeral2} +\expandafter{\romannumeral3} +\expandafter{\romannumeral4} +\expandafter{\romannumeral5})/5\\ \expandafter{\romannumeral1}. Divide the value of the Range of High Quality Score among various users by their mean\\ \expandafter{\romannumeral2}. Divide the value of the Range of Reliability Score among various users by their mean\\ \expandafter{\romannumeral3}. Divide the value of the Range of Identity Score among various users by their mean\\ \expandafter{\romannumeral4}. Divide the value of the Range of Cooperation Score among various users by their mean\\ \expandafter{\romannumeral5}. Divide the value of the Range of Social Awareness Score among various users by their mean\end{tabular} \\ \midrule
\multicolumn{1}{l|}{\textbf{Overall Score}} & \begin{tabular}[c]{@{}l@{}}Evaluate the overall performance\\ given all ability-specific scores.\end{tabular} & Ability-specific scoring rubrics (1-5)  \\ \bottomrule
\end{tabular}%
}
\caption{Summary of the evaluation taxonomy, descriptions of abilities, and evaluation methods in \textsc{Concept}. The LLM-based evaluator is used for ability-specific scoring, whereas computational metrics are used for the rest.}
\label{tab:summary_full}
\end{table*}

\clearpage
\section{Checklist}

\begin{enumerate}

\item For all authors...
\begin{enumerate}
  \item Do the main claims made in the abstract and introduction accurately reflect the paper's contributions and scope?
    \answerYes{See Section \ref{introthis} and our Abstract.}
  \item Did you describe the limitations of your work?
    \answerYes{See Appendix \ref{limitation}.}
  \item Did you discuss any potential negative societal impacts of your work?
    \answerNA{Instead of introducing new CRS models with potential societal ramifications, we assess the practical applications of existing ones.}
  \item Have you read the ethics review guidelines and ensured that your paper conforms to them?
    \answerYes{We ensure that this paper does not involve any ethical issues.}
\end{enumerate}

\item If you are including theoretical results...
\begin{enumerate}
  \item Did you state the full set of assumptions of all theoretical results?
    \answerNA{No theoretical result is involved}
	\item Did you include complete proofs of all theoretical results?
    \answerNA{No theoretical result is involved}
\end{enumerate}

\item If you ran experiments (e.g. for benchmarks)...
\begin{enumerate}
  \item Did you include the code, data, and instructions needed to reproduce the main experimental results (either in the supplemental material or as a URL)?
    \answerYes{\url{https://github.com/huangzichun/Concept4CRS}}
  \item Did you specify all the training details (e.g., data splits, hyperparameters, how they were chosen)?
    \answerYes{We do not involve any model training. However, we offer model evaluation details in Appendix \ref{detailsin}.}
	\item Did you report error bars (e.g., with respect to the random seed after running experiments multiple times)?
    \answerYes{We present the average results across various age groups of users.}
	\item Did you include the total amount of compute and the type of resources used (e.g., type of GPUs, internal cluster, or cloud provider)?
    \answerYes{See Appendix \ref{detailsin}.}
\end{enumerate}

\item If you are using existing assets (e.g., code, data, models) or curating/releasing new assets...
\begin{enumerate}
  \item If your work uses existing assets, did you cite the creators?
    \answerNA{The dataset utilized in this research was dynamically generated in conjunction with our novel evaluation protocol.}
  \item Did you mention the license of the assets?
    \answerNA{}
  \item Did you include any new assets either in the supplemental material or as a URL?
    \answerYes{\url{https://github.com/huangzichun/Concept4CRS}}
  \item Did you discuss whether and how consent was obtained from people whose data you're using/curating?
    \answerNA{The dataset utilized in this research was dynamically generated in conjunction with our novel evaluation protocol.}
  \item Did you discuss whether the data you are using/curating contains personally identifiable information or offensive content?
    \answerYes{The dataset utilized in this research was dynamically generated in conjunction with our novel evaluation protocol. There is no personally identifiable information. We carefully prompt the aligned LLMs to avoid any offensive content.}
\end{enumerate}

\item If you used crowdsourcing or conducted research with human subjects...
\begin{enumerate}
  \item Did you include the full text of instructions given to participants and screenshots, if applicable?
    \answerNA{The dataset utilized in this research was dynamically generated by LLMs}
  \item Did you describe any potential participant risks, with links to Institutional Review Board (IRB) approvals, if applicable?
    \answerNA{The dataset utilized in this research was dynamically generated by LLMs}
  \item Did you include the estimated hourly wage paid to participants and the total amount spent on participant compensation?
    \answerNA{The dataset utilized in this research was dynamically generated by LLMs}
\end{enumerate}

\end{enumerate}


\section{Datasheet}
To clarify, our contribution lies in the evaluation protocol, not the dataset. The dataset is generated dynamically alongside the execution of the protocol. To prevent any confusion regarding the generated dataset, we also included a datasheet for reference. Note that this datasheet adheres to the guidelines outlined in~\citet{gebru2021datasheets}.

\begin{enumerate}
    \item Motivation
    \begin{enumerate}
    \item For what purpose was the dataset created? \\
    This dataset was created to evaluate and analyze the strengths, weaknesses, and potential risks of off-the-shelf CRS models. 
    \item Who created the dataset and on behalf of which entity? \\
    This dataset was generated with the aid of LLMs and CRS models, with the primary code developed by Peixin Qin and Chen Huang. 
    \item Who funded the creation of the dataset? \\
    \answerNA{}
    \item Any other Comments? \\
    \answerNo{}
    \end{enumerate}
    \item Composition
    \begin{enumerate}
        \item What do the instances that comprise the dataset represent? \\
        A conversation between a user simulator and a CRS model.
        \item How many instances are there in total? \\
        There are 6720 conversations in total.
        \item Does the dataset contain all possible instances or is it a sample (not necessarily random) of instances from a larger set? \\
        It contains all possible instances.
        \item What data does each instance consist of? \\
        It comprises a multi-turn conversation between a user and a CRS model, along with the user's persona, age, preferences, and the specific name of CRS model employed.
        \item Is there a label or target associated with each instance? \\
        Each conversation is associated a targeted movie set as the recommendation ground truth.
        \item Is any information missing from individual instances? \\
        \answerNo{}
        \item Are relationships between individual instances made explicit? \\
        \answerNo{}
        \item Are there recommended data splits? \\
        \answerNo{All gathered data is intended for CRS model evaluation.}
        \item Are there any errors, sources of noise, or redundancies in the dataset? \\
        \answerNA{}
        \item Is the dataset self-contained, or does it link to or otherwise rely on external resources (e.g., websites, tweets, other datasets)? \\
        \answerYes{We utilize established movie recommendation datasets as the source for user preference candidates.}
        \item Does the dataset contain data that might be considered confidential (e.g., data that is protected by legal privilege or by doctor-patient confidentiality, data that includes the content of individuals' non-public communications)? \\
        \answerNo{}
        \item Does the dataset contain data that, if viewed directly, might be offensive, insulting, threatening, or might otherwise cause anxiety? \\
        \answerNo{We dedicate significant effort to reviewing every conversation.}
        \item Does the dataset relate to people? \\
        \answerYes{We did not involve real human in our dataset. We employed a user simulator.}
        \item Does the dataset identify any subpopulations (e.g., by age, gender)? \\
        \answerYes{We simulated various users with different age and personas.}
        \item Is it possible to identify individuals (i.e., one or more natural persons), either directly or indirectly (i.e., in combination with other data) from the dataset? \\
        \answerNo{All users are simulated.}
        \item Does the dataset contain data that might be considered sensitive in any way (e.g., data that reveals racial or ethnic origins, sexual orientations, religious beliefs, political opinions or union memberships, or locations; financial or health data; biometric or genetic data; forms of government identification, such as social security numbers; criminal history)?  \\
        \answerNo{}
        \item Any other comments? \\
        \answerNo{}
    \end{enumerate}
    \item Collection Process
    \begin{enumerate}
        \item How was the data associated with each instance acquired? \\
        Every conversation between a user and the CRS is represented as a single instance in our dataset.
        \item What mechanisms or procedures were used to collect the data (e.g., hardware apparatus or sensor, manual human curation, software program, software API)? \\
        We resort to the LLMs. See Section \ref{evaluation}.
        \item If the dataset is a sample from a larger set, what was the sampling strategy (e.g., deterministic, probabilistic with specific sampling probabilities)? \\
        \answerNA{}
        \item Who was involved in the data collection process (e.g., students, crowdworkers, contractors), and how were they compensated (e.g., how much were crowdworkers paid)? \\
        Only authors are involved in the data collection.
        \item Over what timeframe was the data collected? \\
        All data are collected in 2024-01. 
        \item Were any ethical review processes conducted (e.g., by an institutional review board)? \\
        \answerNo{No ethical review processes were conducted with respect to the collection and annotation of this data.}
        \item Does the dataset relate to people? \\
        \answerYes{We did not involve real human in our dataset. We employed a user simulator.}
        \item Did you collect the data from the individuals in question directly, or obtain it via third parties or other sources (e.g., websites)? \\
        \answerNo{We collected the data by sorting to the LLMs.} 
        \item Were the individuals in question notified about the data collection? \\
        \answerNA{} 
        \item Did the individuals in question consent to the collection and use of their data? \\
        \answerNA{} 
        \item If consent was obtained, were the consenting individuals provided with a mechanism to revoke their consent in the future or for certain uses? \\
        \answerNA{} 
        \item Has an analysis of the potential impact of the dataset and its use on data subjects (e.g., a data protection impact analysis) been conducted? \\
        \answerNA{}
        \item Any other comments? \\
        \answerNo{}
    \end{enumerate}
    \item Preprocessing, Cleaning and Labeling
    \begin{enumerate}
        \item Was any preprocessing/cleaning/labeling of the data done (e.g., discretization or bucketing, tokenization, part-of-speech tagging, SIFT feature extraction, removal of instances, processing of missing values)? \\
        \answerNo{No preprocessing/cleaning/labeling is needed.}
        \item Was the "raw" data saved in addition to the preprocessed/cleaned/labeled data (e.g., to support unanticipated future uses)? \\
        \answerNA{The dataset userd in the paper is the raw dataset itself. We release the raw data on the Github}.
        \item Is the software used to preprocess/clean/label the instances available? \\
        \answerNA{}
        \item Any other comments? \\
        \answerNo{}
    \end{enumerate}
    \item Uses
    \begin{enumerate}
        \item Has the dataset been used for any tasks already? \\
        \answerNo{}
        \item Is there a repository that links to any or all papers or systems that use the dataset? \\
        \answerNo{}
        \item What (other) tasks could the dataset be used for? \\
        This dataset can help research ways to improve the persuasiveness and honesty of CRS explanations.
        \item Is there anything about the composition of the dataset or the way it was collected and preprocessed/cleaned/labeled that might impact future uses? \\
        \answerNA{No preprocessing/cleaning/labeling}
        \item Are there tasks for which the dataset should not be used? \\
        The usage of this dataset should be limited to the scope of CRS.
        \item Any other comments? \\
        \answerNo{}
    \end{enumerate}
    \item Distribution
    \begin{enumerate}
        \item Will the dataset be distributed to third parties outside of the entity (e.g., company, institution, organization) on behalf of which the dataset was created? \\
        \answerYes{We expect other websites re-distribute our dataset.}
        \item How will the dataset be distributed (e.g., tarball on website, API, GitHub)? \\
        The dataset could be accessed on a GitHub webpage.
        \item When will the dataset be distributed? \\
        The dataset is already released to the Github.
        \item Will the dataset be distributed under a copyright or other intellectual property (IP) license, and/or under applicable terms of use (ToU)? \\
        \answerNA{The dataset is generated dynamically alongside the execution of the protocol. No license is needed}.
        \item Have any third parties imposed IP-based or other restrictions on the data associated with the instances? \\
        \answerNo{}
        \item Do any export controls or other regulatory restrictions apply to the dataset or to individual instances? \\
        \answerNo{}
        \item Any other comments? \\
        \answerNo{}
    \end{enumerate}
    \item Maintenance
    \begin{enumerate}
        \item Who is supporting/hosting/maintaining the dataset? \\
        Chen Huang will be responsible for maintaining the dataset. 
        \item How can the owner/curator/manager of the dataset be contacted (e.g., email address)? \\
        E-mail addresses are at the top of the paper.
        \item Is there an erratum? \\
        \answerNo{When errors are discerned, we will announce erratum on our website}.
        \item Will the dataset be updated (e.g., to correct labeling errors, add new instances, delete instances')? \\
        \answerYes{We hope to expand this dataset with more CRS models and recommendation scenarios.}
        \item If the dataset relates to people, are there applicable limits on the retention of the data associated with the instances (e.g., were individuals in question told that their data would be retained for a fixed period of time and then deleted)? \\
        \answerNA{}
        \item Will older versions of the dataset continue to be supported/hosted/maintained? \\
        \answerYes{}
        \item If others want to extend/augment/build on/contribute to the dataset, is there a mechanism for them to do so? \\
        \answerNo{}
        \item Any other comments? \\
        \answerNo{}
    \end{enumerate}
\end{enumerate}

\end{document}